\documentclass[journal]{IEEEtran}
\usepackage{ifpdf}

\usepackage{cite}

\ifCLASSINFOpdf
  \usepackage[pdftex]{graphicx}
\else
\fi
\usepackage{multirow, boldline}
\usepackage{amsfonts}
\usepackage{booktabs}
\usepackage{rotating}
\usepackage{tabularx}
\usepackage{xcolor}

\usepackage{amsmath}
\usepackage{algorithmic}

\usepackage{array}

\usepackage[caption=false,font=footnotesize]{subfig}
\begin{document}
\title{Relational Deep Feature Learning \\for Heterogeneous Face Recognition}
%
%
%

\author{MyeongAh Cho,
        Taeoh Kim,
        Ig-Jae Kim,
        Kyungjae Lee,
        and~Sangyoun Lee,~\IEEEmembership{Member,~IEEE}
\thanks{This research was supported by R\&D program for Advanced Integrated-intelligence for Identification (AIID) through the National Research Foundation of KOREA(NRF) funded by Ministry of Science and ICT (NRF-2018M3E3A1057289)
	
	M.Cho, T.Kim and S.Lee are with the School of Electrical and Electronic Engineering,
	Yonsei University, Seoul, South Korea (e-mail: maycho0305@yonsei.ac.kr;
	kto@yonsei.ac.kr; syleee@yonsei.ac.kr).}
\thanks{I.Kim is with the Center for Imaging Media Research, Korea Institute of Science and Technology, Seoul, South Korea (e-mail: drjay@kist.re.kr).}
\thanks{K.Lee is with the Department of Computer Science, Yongin University, Yongin, South Korea (kjlee@yiu.ac.kr).}} 

 
%

\markboth{IEEE TRANSACTIONS ON INFORMATION FORENSICS AND SECURITY,~Vol.~xx, No.~x, x~xxxx}%
{MyeongAh Cho \MakeLowercase{\textit{et al.}}}
%



\maketitle

\begin{abstract}

Heterogeneous Face Recognition (HFR) is a task that matches faces across two different domains such as visible light (VIS), near-infrared (NIR), or the sketch domain. Due to the lack of databases, HFR methods usually exploit the pre-trained features on a large-scale visual database that contain general facial information. However, these pre-trained features cause performance degradation due to the texture discrepancy with the visual domain. With this motivation, we propose a graph-structured module called Relational Graph Module (RGM) that extracts global relational information in addition to general facial features. Because each identity’s relational information between intra-facial parts is similar in any modality, the modeling relationship between features can help cross-domain matching. Through the RGM, relation propagation diminishes texture dependency without losing its advantages from the pre-trained features. Furthermore, the RGM captures global facial geometrics from locally correlated convolutional features to identify long-range relationships. In addition, we propose a Node Attention Unit (NAU) that performs node-wise recalibration to concentrate on the more informative nodes arising from relation-based propagation. Furthermore, we suggest a novel conditional-margin loss function ($C$-softmax) for the efficient projection learning of the embedding vector in HFR.

The proposed method outperforms other state-of-the-art methods on five HFR databases. Furthermore, we demonstrate performance improvement on three backbones because our module can be plugged into any pre-trained face recognition backbone to overcome the limitations of a small HFR database
\end{abstract}


\ifCLASSOPTIONpeerreview
\begin{center} \bfseries EDICS Category: 3-BBND \end{center}
\fi
%
\IEEEpeerreviewmaketitle

\section{Introduction} \label{Introduction}
%
%
%
%
\IEEEPARstart{F}{ace recognition}, a task that aims to match facial images of the same person, has developed rapidly with the advent of deep learning. The features extracted through the multiple-hidden layers of deep convolutional neural network (DCNNs) contain representative information that is used to distinguish an individual\cite{wang2018deep}. However, when recognizing a face via representative features, variations such as pose, illumination, or facial expression create difficulties\cite{sengupta2016frontal, wang2018face, zangeneh2020low}. Unlike general face recognition within the visible spectrum, the Heterogeneous Face Recognition (HFR) aims to match faces across different domains such as visible light (VIS), near-infrared (NIR), or the sketch domain\cite{he2017learning, bi2019multi}. Face recognition over different domains is important, since NIR images acquired with infrared cameras contain more useful information when visible light is lacking, while sketch-to-photo matching is important in law enforcement for rapidly identifying suspects\cite{wang2018deep}. As such, HFR can be a practical application for biometric security control or surveillance cameras under low-light scenarios\cite{ouyang2016survey}.

\begin{figure*}[!t]
	\centering
	\subfloat[CASIA NIR-VIS 2.0]{\includegraphics[width=0.25\textwidth]{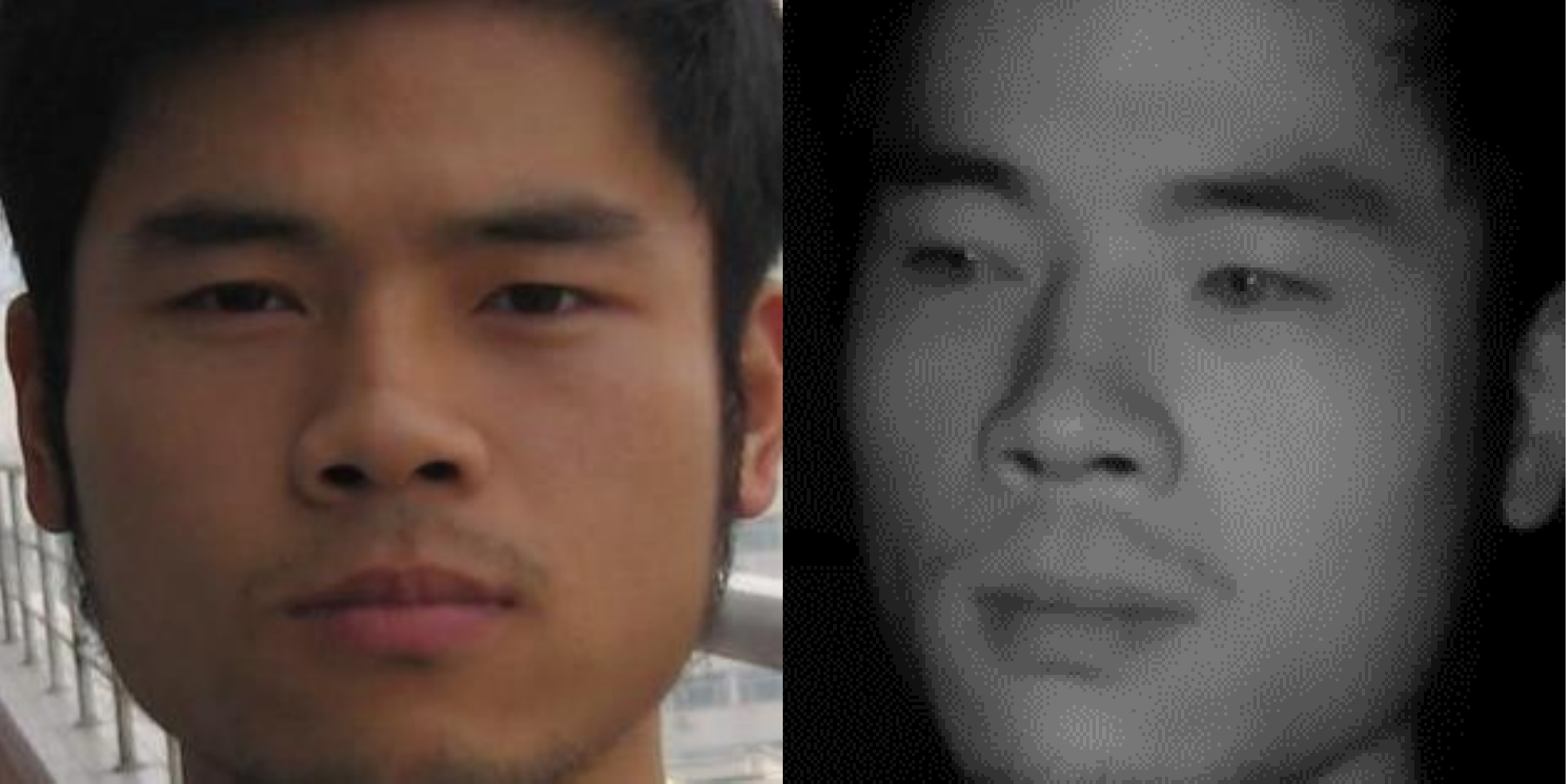}%
		\label{f2_a}}
	\hfil
	\subfloat[IIIT-D Sketch]{\includegraphics[width=0.25\textwidth]{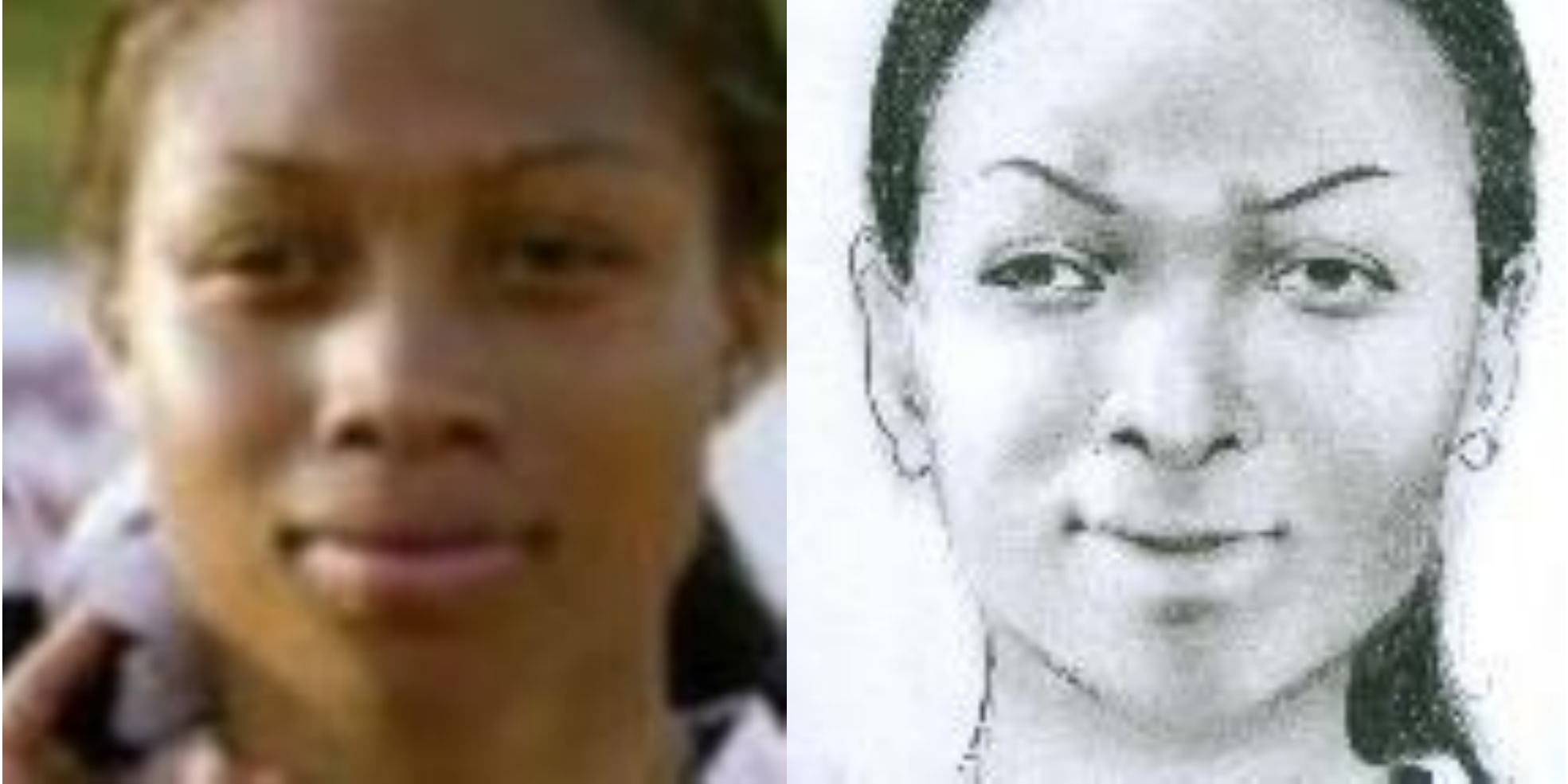}%
		\label{f2_b}}
	\hfil
	\subfloat[BUAA Vis-Nir]{\includegraphics[width=0.25\textwidth]{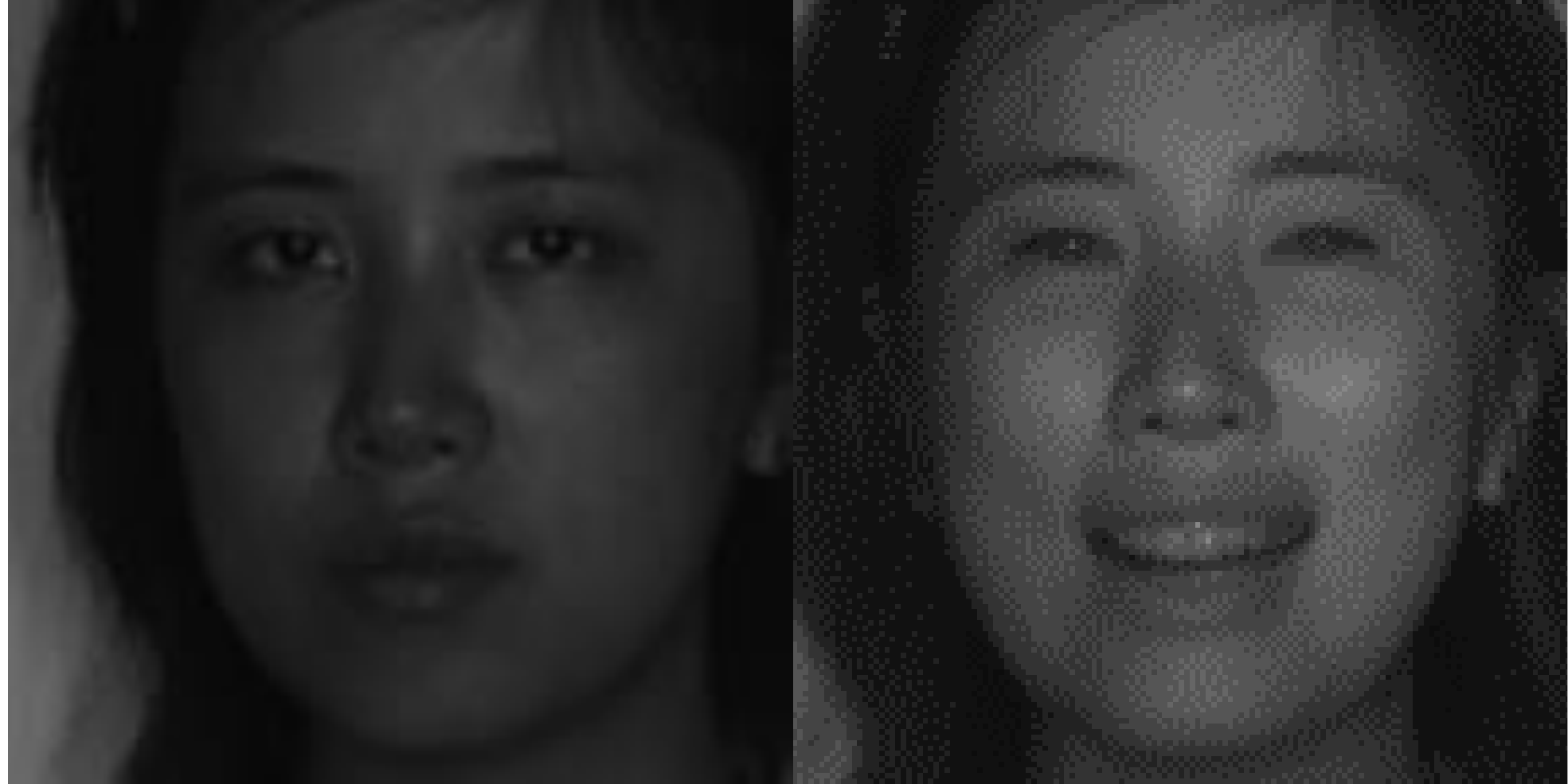}%
		\label{f2_c}}
	\hfil
	\caption{Examples of CASIA NIR-VIS 2.0\cite{li2013casia}, IIIT-D Sketch\cite{bhatt2012memetic} and BUAA Vis-Nir\cite{huang2012buaa} databases. In each pair, the left and right sides are VIS and non-VIS data samples, respectively, and these pairs show large domain discrepancies.}
	\label{f2}
\vspace{-2mm}
\end{figure*}

HFR has several challenging issues, the biggest of which is the large gap between data domains. When HFR is performed with a face recognition network pre-trained on VIS face images, accuracy is significantly reduced. This is because the difference between the distributions of VIS and non-VIS data is large. Therefore, we need to reduce the domain gap through either learning domain-invariant features or common space projection methods. Another issue is the lack of HFR databases. Deep learning-based face recognition networks are usually trained with large-scale visual database such as MS-Celeb-1M\cite{guo2016ms} which consists of 10 million images with 85 thousand identities, or MegaFace\cite{kemelmacher2016megaface}, with 4.7 million images. By comparison, the typical HFR database has a small number of images and subjects, which causes overfitting in a deep network and makes learning a general feature difficult. Therefore, most HFR tasks are fine-tuned using a backbone which is pre-trained on a large visual database.

To solve these problems, several works \cite{bi2019multi, song2018adversarial, zhang2018tv} use image synthesis method to transform input images from non-VIS domain to VIS domains and recognize faces in the same domain feature space. Although this approach may create a similar domain by the the data transform, it is difficult to generate good-quality transform images with a small amount of data; this greatly impacts performance and the approach does not reduce the gap between domain properties. Other studies train the network to learn NIR-VIS-invariant features by using the Wasserstein distance\cite{he2018wasserstein}, variational formulation\cite{wu2018coupled}, a triplet loss function\cite{liu2016transferring}, or domain-specific module \cite{deng2019residual}. These domain-invariant approaches force the network to reduce the domain gap implicitly, which makes learning and designing a network challenging. Therefore, we propose a graph-structured module that reduces the fundamental differences in heterogeneous domain characteristics by extracting global relational information using general facial features from the large-scale visual dataset.

For many computer vision tasks, the relational information within the image or video is important, in the same way that human visual processing can easily perform recognition by capturing relations. Since each identities’ relational information between intra facial parts are very similar in any modality, it is suitable for reducing the gap between domains in HFR. With our proposed Relational Graph Module (RGM), each component of the face is embedded to a node vector and edges are computed by modeling the relationships among nodes. Through graph propagation with generated nodes and edges, we create a relational node vector containing the overall relationship nodes; and perform node-wise recalibration through the correlation information for these nodes with Node Attention Unit (NAU). Also, we suggest a conditional-margin loss function ($C$-softmax) to learn with an efficient space margin between inter-classes when data from two different domains is projected into one latent space. 

In this paper, our main contributions are as follows:
\begin{itemize}
	\setlength{\itemsep}{0.5em}
	\item We propose a graph-structured module, RGM, to reduce the fundamental domain gap by modeling face components as node vectors and relational information edges. We also perform a recalibration by considering global node correlation via NAU. 
	\item In order to project features from different domains into common latent space efficiently, we suggest a $C$-softmax that uses the inter-class margin conditionally. 
	\item The proposed module can overcome the limitation of HFR databases by plugging to a general feature extractor; we experimentally demonstrate superior performance for three different backbones and five HFR databases.
\end{itemize}
The organization of the paper is as follows. In Section~\ref{related works}, we introduce three different approaches to HFR and  briefly describe some relation capturing methods. Then, in Section~\ref{proposed method}, we begin by presenting a preliminary version of this work, Relation Module (RM)\cite{myeong2019rm}, and explain our proposed RGM, NAU, and $C$-softmax approaches for HFR. Next, in Section~\ref{Experimental results}, the experimental results and related discussions are provided, prior to conclusion in Section~\ref{conclusion}.

\begin{figure*}[!t]
	\centering
	\subfloat[Bilinear module]{\includegraphics[width=0.5\textwidth]{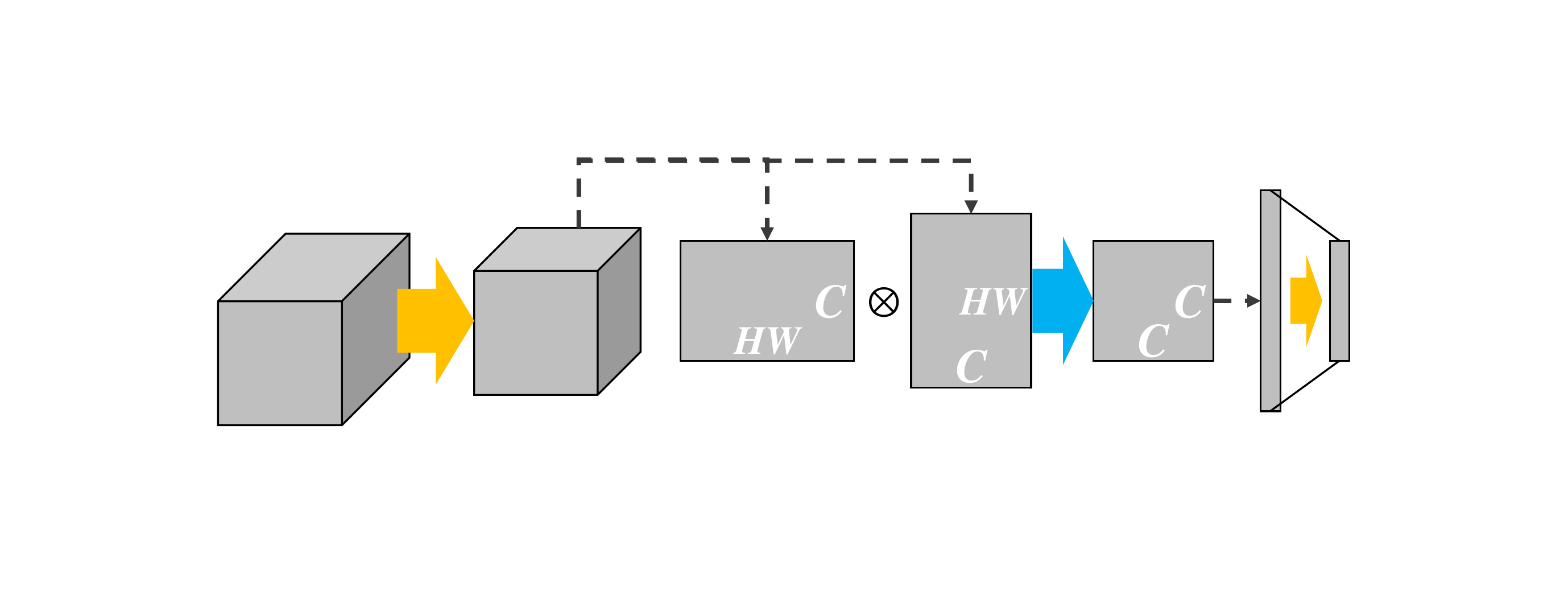}%
		\label{f3_1}}
	\hfil
	\subfloat[Double Attention module]{\includegraphics[width=0.5\textwidth]{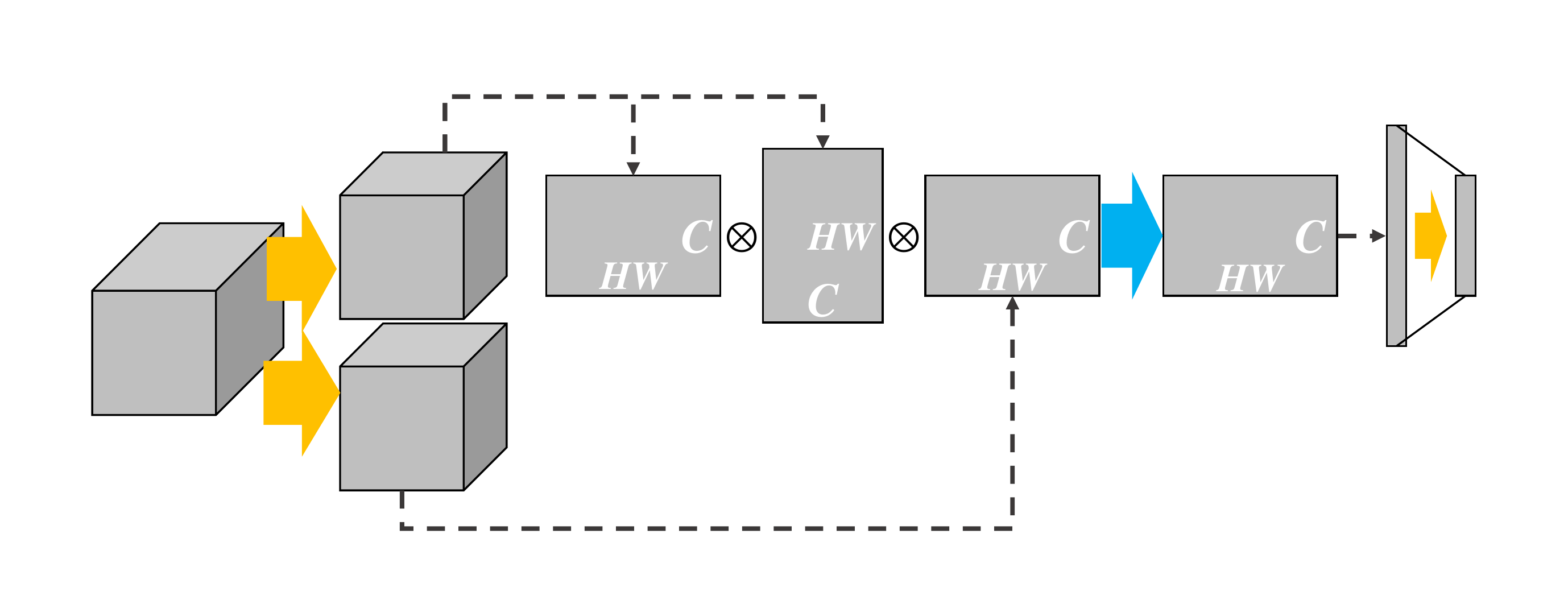}%
		\label{f3_2}}
	\hfil
	\vspace{-0.7\baselineskip}
	\subfloat[Non-Local module]{\includegraphics[width=0.5\textwidth]{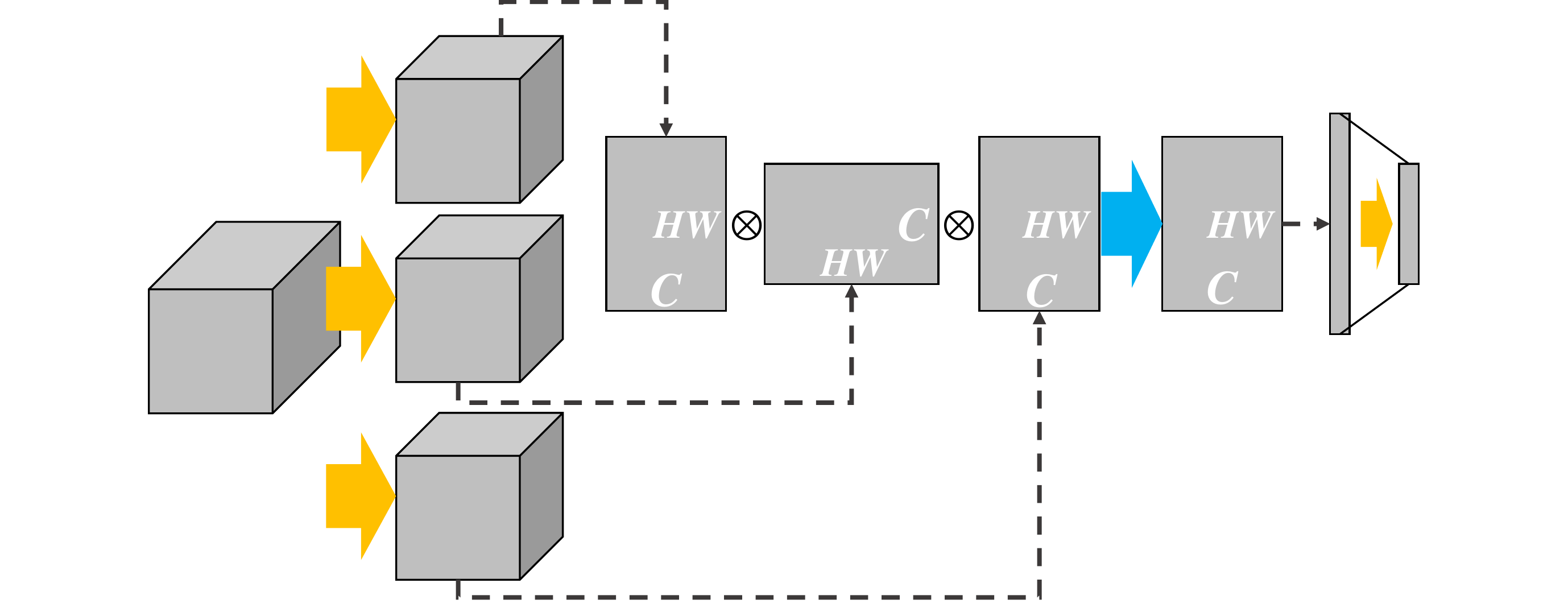}%
		\label{f3_3}}
	\hfil
	\subfloat[Relation Module]{\includegraphics[width=0.5\textwidth]{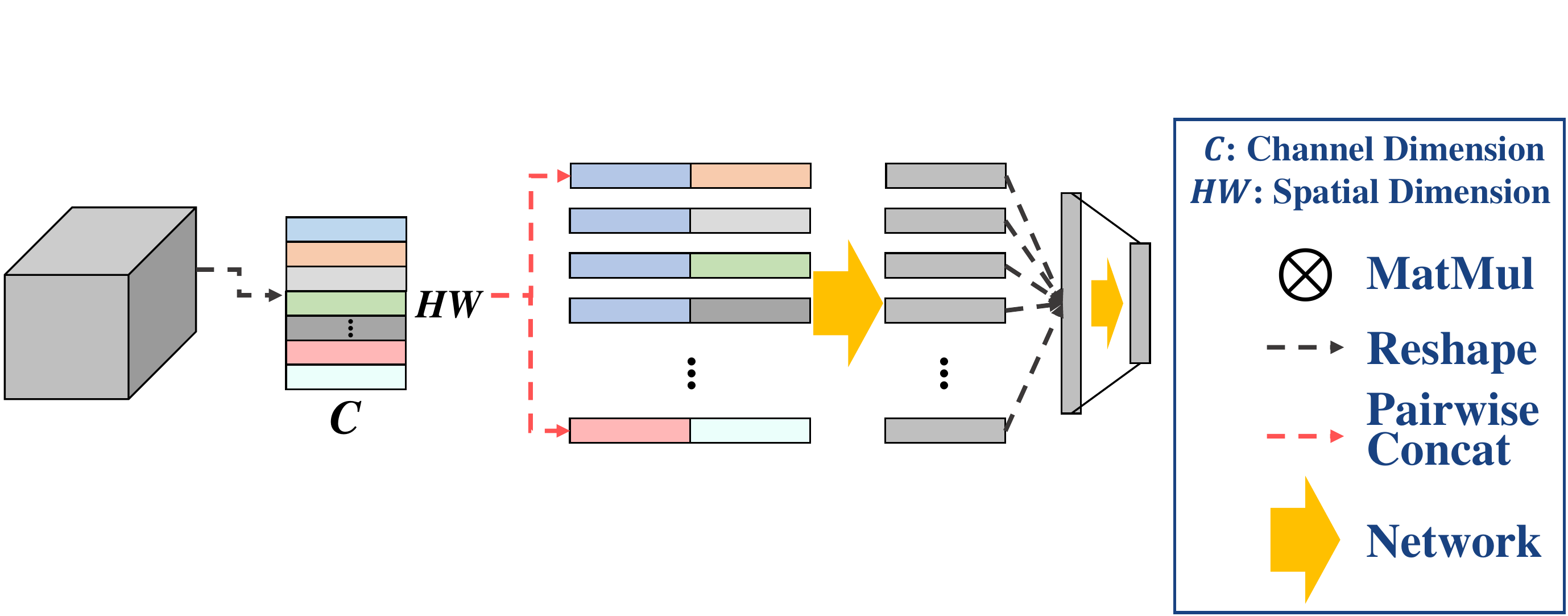}%
		\label{f3_4}}
	\caption{Frameworks of various attentional modules and RM. Each module’s input is a CNN feature map and its output is an $L$-dimensional embedding vector. Module (a) is from \cite{chowdhury2016one} and modules (b), (c), and (d) are modified from the original structure\cite{chen20182, wang2018non, santoro2017simple}, respectively. For face recognition, the process of reshaping and embedding features through the fully connected layer is added in modules (b), (c), and (d).
	}
	\label{f3}
\end{figure*}

\section{Related Works} \label{related works}
As stated in Section~\ref{Introduction}, the challenge of the HFR task to match identities using conventional face recognition networks despite such domain differences as texture or style. The examples in Figure~\ref{f2} of NIR-to-VIS and Sketch-to-Photo databases illustrate the gaps between the domains which can depend on variations in illumination or on the artist's sketch style. Therefore, methods for reducing domain discrepancy are being studied, and these can be largely divided into projection to common space based methods, image synthesis based methods, and domain-invariant feature based methods. This section summarizes preceding HFR studies and then introduces methods of capturing the relational information within the image that can reduce the fundamental domain difference.

\subsection{Heterogeneous Face Recognition}


\subsubsection{Projection to Common Space Based Method}

Projection based methods involve learning to project features from two different domains into a discriminative common latent space where images with the same identity are close regardless of their domain. Lin and Tang \cite{lin2006inter} proposed a Common Discriminant Feature Extraction (CDFE) algorithm, in which two different domain features simultaneously learn common space to solve the inter-modality problem. With empirical separability and a local consistency regularization objective function, the model learns compact intra-class space and prevents the overfitting problem. Yi and Liao \cite{yi2009partial} suggested matching each partial patch of face images by extracting points, edges, or contours that are similar between domains. Lei \textit{et al.} \cite{lei2009coupled} designed a Coupled Spectral Regression (CPR) method for finding different projective vectors by representing relationships between each images and their embeddings. Different from this coupled method, which learns each domain’s representative vectors separately, Lei \textit{et al.} \cite{lei2012coupled} suggested the technique of learning the projection from both domains. Since target data neighbors should correspond to source data neighbors, Shao \textit{et al.} \cite{shao2014generalized} matched projected target and source data by using a reconstruction coefficient matrix, Z, in the common space. 

With deep neural networks (DNNs) showing great improvement in face recognition performance, Sarfraz and Stiefelhagen \cite{sarfraz2015deep} used a deep perceptual mapping (DPM) method in which DNNs learn projection of visual and thermal features together. In \cite{reale2016seeing}, Reale \textit{et al.} used coupled NIR and VIS CNNs, initializing them with a pre-trained face recognition network to extract global features. Wu \textit{et al.} \cite{wu2018coupled} proposed a Coupled Deep Learning (CDL) method with relevance constraints as a regularizer and a cross-modal ranking objective function. However, these methods are difficult to train because they require extraction of domain-specific features with a small database.

\subsubsection{Image Synthesis Based Method}
Image synthesis based methods transform face images from one domain into the other so as to perform recognition in the same modality. Liu \textit{et al.} \cite{liu2005nonlinear} proposed a pseudo-sketch synthesis method which divides a photo image into a fixed number of patches and reconstructs each patch as a corresponding sketch patch. This patch-based strategy preserves local geometry while transforming the photo image into a sketch-style image. In \cite {wang2008face}, Wang \textit{et al.} proposed a multi-scale Markov network, conducting brief propagation to transform multi-scale patches. Recently, with the widespread development of generative adversarial networks (GANs) \cite{goodfellow2014generative}, many studies have focused on generating visual face images from non-visual ones\cite{zhang2018tv}, \cite{bi2019multi}. In \cite{song2018adversarial} Song \textit{et al.} transformed NIR face images to VIS face images by pixel space adversarial learning using CycleGAN\cite{zhu2017unpaired} and feature space adversarial learning with a variance discrepancy loss function. 

These methods of transforming an image from one domain to another can be effective for visually similar domain, but do not fundamentally address the modality discrepancy that the data exhibits. In addition, due to issue of small amounts of unpaired HFR data, GAN-based methods struggle to create good-quality images, which affects performance.

\subsubsection{Domain-invariant Feature Based Method}
Another approach is to use a feature extractor to reduce domain discrepancies and enable learning of domain-invariant features. Since the NIR-to-VIS face recognition task is heavily influenced by the light source in each image, Liu \textit{et al.} \cite{liu2012heterogeneous} used differential-based band-pass image filters, relying only on variation patterns of skin properties. Lui \textit{et al.} \cite{liu2016transferring} also proposed a TRIVET loss function which applies Triplet loss \cite{schroff2015facenet} to cross-domain sampling to reduce the domain gap. He \textit{et al.} \cite{he2017learning} used a division approach, using two orthogonal spaces to represent invariant identity and light source information. The Wasserstein CNN in \cite{he2018wasserstein} is also divided into an NIR-VIS shared layer and a specific layer. The shared layer is designed to learn domain-invariant features by minimizing the Wasserstein distance between different domain data. Based on this approach, the DVR method \cite{wu2019disentangled} is proposed to disentangle the cross-domain facial representations with the identity information and within-person variations. 

Several studies \cite{klare2012heterogeneous}, \cite{peng2016graphical}, \cite{peng2019sparse} used relational representations that allow projection heterogeneous data into a common space. In \cite{klare2012heterogeneous}, Klare and Jain proposed a random prototype subspace framework to define prototype representations and learn a subspace projection matrix with kernel similarities among face patches. With their G-HFR method \cite{peng2016graphical}, Peng \textit{et al.} employed Markov networks to extract graphical representations. This method finds the $k$ nearest patches of a patch in a probe or gallery image from the representation dataset and linearly combines them to obtain graphical representations. Since finding $k$ nearest patches from the representation process performance score relies heavily on the value of $k$, Peng \textit{et al.} \cite{peng2019sparse} proposed an adaptive spare graphical representation method which considers all possible numbers of related image patches. These methods found relations between representation dataset image patches in randomly selected pairs. Unlike these methods, our proposed RGM extracts domain-invariant features by considering global spatial pair-wise relations. We first apply a deep learning based relation approach to the HFR task with graph structured module.

\begin{figure*}[!t]
	\centering
	\includegraphics[width=\textwidth]{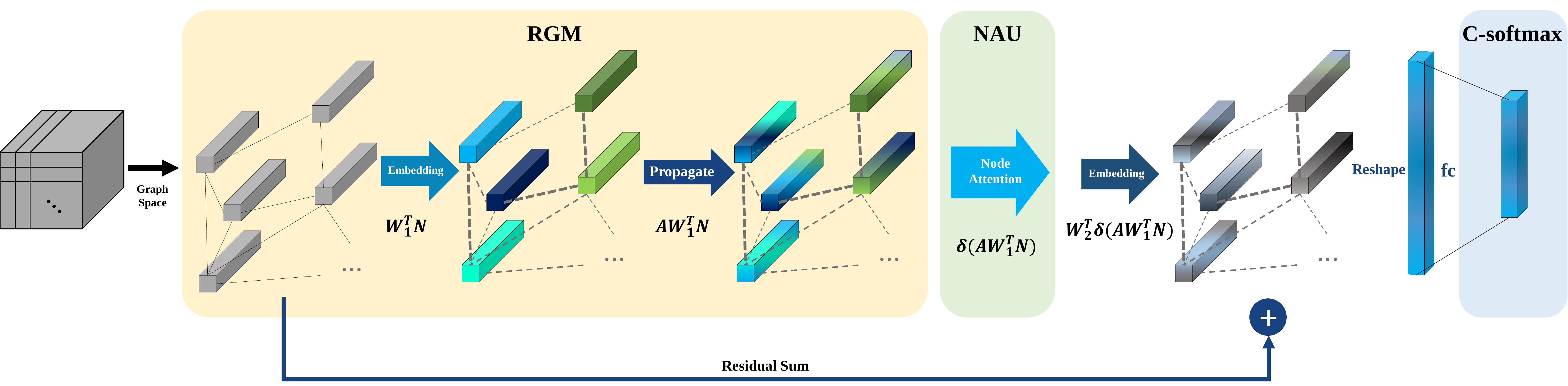}
	\caption{Overall framework of our proposed approach. The feature map, which is an output of the pre-trained backbone, is treated as node vectors of a graph that represent each spatial region, including hair, eyes, mouth and so on. After node-wise embedding, directed pair-wise relation values are calculated. From the directed adjacency matrix $A$, relations between nodes are propagated in RGM and recalibrated via NAU. After residual summation, whole relation vectors are reshaped and embedded into a representative vector which go through the $C$-softmax layer.}
	\label{f4}
\end{figure*}

\subsection{Relation Capturing}
In many computer vision tasks such as image classification, video recognition, and so on, it is important to understand the relationships within images or videos. However, simply operating multiple neural network layers often fails to identify long-range relationships as human visual systems do. With CNN bring significant improvements in computer vision, there are many studies underway to extract relational information using the local connectivity and multi-layer structure of CNN.

In \cite{lin2015bilinear}, Lin \textit{et al.} captured local pair-wise feature interaction to solve the fine-grained categorization task, in which visual differences are small between classes and can easily be overwhelmed by other factors such as view point or pose. The features from two streams of CNN termed Bilinear CNN or B-CNN are multiplied using an outer product to capture partial feature correlation. Since the face recognition task can be seen as a subarea of fine-grained recognition, Chowdhury \textit{et al.} \cite{chowdhury2016one} applied this bilinear model to face recognition tasks with symmetric B-CNN (see Figure~\ref{f3}(a)). Chen \textit{et al.} \cite{chen20182} captured relational information with a double attention block consisting of bilinear pooling attention and feature distribution attention (Figure~\ref{f3}(b)). The Non-local block \cite{wang2018non} was proposed to operate a weighted sum of all features at each position, showing outperformance in video recognition tasks (Figure~\ref{f3}(c)). To solve the Visual Q\&A task, which requires relation reasoning information, Santoro \textit{et al.} \cite{santoro2017simple} introduced a Relation Network that captures all potential relations for object pairs and it was applied to other tasks \cite{sun2018actor, myeong2019rm} (Figure~\ref{f3}(d)). 

Recently, graph-based methods have proven effective in relation capturing\cite{zhou2018graph}. While traditional graph analysis has usually relied on hand crafted features, Graph Neural Networks (GNN) can learn nodes or edges update by propagating each layer’s weights. Kipf and Welling \cite{kipf2016semi} proposed a spectral method using a Graph Convolutional Networks (GCN) which inputs graph-structured data and uses multiple hidden layers to learn graph structures and node features. Wang and Gupta \cite{wang2018videos} applied GCNs in action recognition to understand appearance and temporal functional relationships, while Chen \textit{et al.}\cite{chen2019graph} proposed a GloRe module that projects coordinate space features into interaction space to extract relation-aware features, boosting the performance of 2D and 3D CNNs on semantic segmentation, image recognition, and so on. As such, the graph-structured networks are effective for most computer vision tasks where relational information is important. In particular, since the HFR task involves small differences between classes and large within class discrepancies, relation information for faces plays an important role in representing each identity. Compared to Attentional Modules \cite{lin2015bilinear,chen20182, wang2018non}, graph-based modules better capture relations, thereby reducing the fundamental domain gap in HFR. We compare existing attentional module-based approaches \cite{lin2015bilinear}, \cite{wang2018non}, \cite{chen20182} as well as graph-based module \cite{chen2019graph} to our module in Section~\ref{relational method}.

\begin{figure*}[t]
	\centering
	\includegraphics[width=0.8\textwidth]{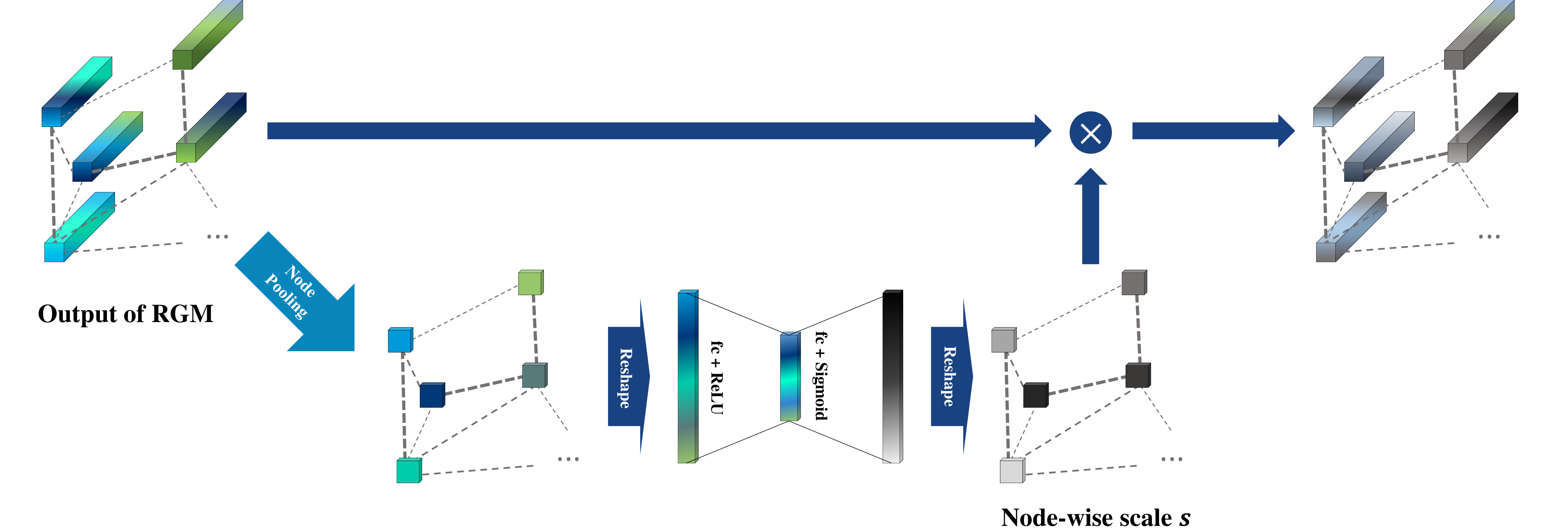}
	\caption{Architecture of the NAU. After RGM propagation, because each node loses spatial correlations, we squeeze nodes point-wisely along embedding dimensions (node pooling). After that, inter-node reasoning is performed with fully connected layers to produce node-wise scale values $\boldsymbol{s}$ (Equation \ref{e4}). These values represent node-wise importance which multiplied to the propagated nodes.}
	\label{f5}

\end{figure*}

\section{Proposed method} \label{proposed method}


In this section, we firstly present our preliminary version of this work called RM \cite{myeong2019rm} and introduce RGM to model and propagate relationships of face components. Secondly, we describe NAU which helps to focus on global node correlation. Last, we introduce a $C$-softmax, loss function with conditional margin. RGM and NAU are add-on module which can be plugged into any pre-trained face recognition backbones. We experimentally quantify the performance improvement of proposed modules in three different backbones and over five different heterogeneous databases in Section~\ref{Experimental results}.

\subsection{Relation Module}
When an NIR or VIS face image is input to a single face recognition network pre-trained with large-scale visual face images, the network cannot perform well because of the domain discrepancy. In addition, HFR databases are mostly unpaired and consist of much smaller numbers of images than large-scale database such as MS-Celeb-1M\cite{guo2016ms} and CASIA-WebFace\cite{yi2014learning}, so it is difficult to fine-tune the pre-trained deep networks. To solve this problem, the RM concentrates on relationships of pair-wise face component that less depend on domain information (see Figure~\ref{f3}(d)). The RM is plugged in at the end of the backbone's convolutional layer and takes an input as a feature map which is the output of the last convolutional layer. This feature map’s spatial-wise vector represents the face component, depending on the CNN's local connectivity characteristics. From these $N\times N$ number of feature vectors, RM extracts the relationships between component pair. Since this pair-wise relationship is independent of ordering, the total number of combinations is $N^{2}\times \left ( N^{2}+1 \right )/2$ and an $L$-dimensional relation vector is extracted from each pair (we use $L$ = 64). 
\begin{gather}
\boldsymbol{RM}\left ( \boldsymbol{f}_{i,j} \right ) = g_{\theta }\left ( \boldsymbol{f}_{i}, \boldsymbol{f}_{j} \right )_{i,j=1,..., N^{2}}
\label{e1}
\end{gather}
In Equation~\ref{e1}, $g_{\theta }\left (\cdot \right )$ is the relation extracting function with shared learnable weight $\theta$ and $\boldsymbol{f}_{i}$, $\boldsymbol{f}_{j}$ are the input feature vectors. The relation vector $ \boldsymbol{RM}\left(\boldsymbol{f}_{i,j}\right)$ represents the relationship between two parts of the face such as the lips-to-nose or eye-to-eye relationship (e.g., distance, ratio or similarity) within a face. For computation, we concatenate two vectors in each combination and embed it into relation vector by shared fully connected layer. These computed relation vectors are reshaped and embedded into one embedding vector with a fully connected layer. 

This process does not need to define actual relationships explicitly but simply looks at all combinations of patches and infers the relationship implicitly. Simply adding the RM can reduce intra-class variation and enlarge the inter-class space by using relational information.

\subsection{Relational Graph Module}
As mentioned, the HFR task suffers from a problem of insufficient data and the difficulty of extracting features that reduce the domain gap. Since we confirmed with the RM that relational information in face images contains domain-invariant information, we propose our RGM for more efficient facial relationship modeling. Because RM considers every pair-wise combination and embeds all of them into $L$-dimensional vector, it presents a computational complexity issue with an attendant overfitting risk when training on a small HFR database. Therefore, we propose a method of relation exploration through our graph-structured RGM, which consists of a node vector containing the face component information and edges that capture relationship information between node vectors. Figure~\ref{f4} shows the overall framework.

\subsubsection{Node Embedding}
We first treat the spatial feature vectors extracted through the backbone as initial graph node vectors with dimension~$C$. Then we embed the node vectors into $d$-dimensional vectors using a transform matrix $\boldsymbol{W}_{1} \in \mathbb{R}^{C \times d}$. We experiment with the optimal value of this embedding dimension $d$ in Section~\ref{ablation}.

\subsubsection{Relation Propagation Based on Directed Relation Extraction} \label{RGM}

The feature vectors of the face image from the convolutional layers represent each face component (e.g., eyes, lips, and chin). In the RM, the feature vectors are simply concatenated to extract the relation through the shared fully connected layer. In RGM, after node embedding, we extract the directed edges of each node. Because the components that represent the face are the same for every classes, we generate a fixed number of component nodes rather than selecting nodes (64 nodes are used in this paper).
\begin{gather}
E_{w_{e}}(\boldsymbol{n}_{i}, \boldsymbol{n}_{j})={\boldsymbol{W}_{e}}^{T}\left [ \boldsymbol{n_{i}}, \boldsymbol{n_{j}} \right ]\nonumber\\
E_{i,j} = E_{w_{e}}(\boldsymbol{n}_{i}, \boldsymbol{n}_{j})\nonumber\\
{A}_{i,j}=\sigma(E_{i,j} )
\label{e2}
\end{gather}
In Equation~\ref{e2}, the edge yielding the relationship between two node vectors $n_{i}$ and $n_{j}$ is obtained through the edge function $E_{w_{e}}(\cdot)$. Edge $E_{i,j}$ is a scalar value and is calculated as the weighted sum between node vector elements, where the weight $\boldsymbol{W}_{e}$ is a parameter obtained through learning. ${A}_{i,j}$ has a value in the range [0, 1] through the sigmoid function $\sigma({\cdot})$. 
\begin{gather}
{\boldsymbol{n}_{i}}^{*} = \sum_{k=1}^{N^{2}}{A}_{i,k}\boldsymbol{n}_{k}
\label{e3}
\end{gather}
Then, as shown in Equation~\ref {e3}, each node vector propagates in inter-dependency with all other node vectors through the edges to become a propagated node vector ${\boldsymbol{n}_{i}}^{*}$. Each face component has different relations for each identity and updating the nodes with the relations can concentrate on component relational information rather than visual-domain features such as texture information.

As a point of comparison, Graph Attention Network (GAT)\cite{velivckovic2017graph} adopt self-attention mechanisms by a learnable parameter $\alpha _{vu}=softmax(g(\boldsymbol{a}^{T}\left [ \boldsymbol{W}^{T}\boldsymbol{n}_{v}||\boldsymbol{W}^{T}\boldsymbol{n}_{u} \right ]))$ which only compute for neighbor nodes $u\in\mathcal{N}_v$ where an adjacency matrix($\mathcal{N}_v$) is used to define this neighbor. By updating nodes $\sum_{u\in\mathcal{N}_v} \alpha_{vu} {\bf W}^{T}\boldsymbol{n}_{u}$ within an adjacency matrix, where $g(\cdot)$ is a LeakyReLU activation function and $\boldsymbol{a}$ is a vector of learnable parameters. In the RGM, the adjacency matrix and $\alpha$ are learned simultaneously; also, the RGM uses a sigmoid activation function which looks at each value separately that allows for independent values of relations. Since the relation between the nodes is independent of the relations of other nodes, sigmoid activation is more relevant than Softmax which looks all values interrelated in phase and  computes the sum of values to 1. We experiment with this activation issue in Section~\ref{sigmoid}.

\subsubsection{Node Re-Embedding}

After propagation, we apply the NAU as an activation function $\delta(\cdot)$ (see Figure~\ref{f4}). The NAU is serves as a node-adaptive activation function, this will be described in detail below. After we recalibrate nodes through the NAU, we use the weight matrix $\boldsymbol{W}_{2} \in \mathbb{R}^{d \times C}$ to re-embed the node vectors into the original input dimension and perform residual summation. Since the feature map from the backbone CNN contains general features of the face, we intended to use general feature and relational information together by residual terms \cite{wang2018non, chen2019graph, wang2018videos}. After summation, we concatenate entire node vectors and embed them into the final representative embedding vector for comparison through the fully connected layer.



\subsection{Node Attention Unit} \label{nau}
In SENet\cite{hu2018squeeze}, when the convolutional layer computes a spatial and channel-wise combination in the receptive field, recalibration of the feature is performed to boost the network’s representative power. Inspired by this approach, each node vector that contains relational information is recalibrated through the NAU by considering inter-node correlation. 
\begin{gather}
z_{i} = \frac{1}{C}\sum^{C}\boldsymbol{{n}_{i}}^{*}(c) \nonumber\\[5pt]
\boldsymbol{s} = \sigma (\boldsymbol{{W}_{b}}^{T} \mathrm{ReLU} (\boldsymbol{{W}_{a}}^{T}\boldsymbol{z}) )\nonumber\\
F_{recalib}(s_{i}, \boldsymbol{{n}_{i}})=s_{i}\boldsymbol{{n}_{i}}
\label{e4}
\end{gather}
In Equation~\ref{e4}, each propagated node $\boldsymbol{{n}_{i}}^{*}$ squeezes information through global average pooling to vector $\boldsymbol{z}$. The node-squeezed vector is then aggregated through weights $\boldsymbol{W_{a}}$ and $\boldsymbol{W_{b}}$ to a node-wise scale vector $\boldsymbol{s}$. Then, with recalibration function $F_{recalib}(\cdot)$, each node vector is scaled (see Figure~\ref{f5}). This process yields a recalibration effect according to the global importance of nodes and focusing attention on the characteristic aspect of the identity. In contrast to the SENet, squeeze spatial dimension and recalibrate channels, we squeeze channels and recalibrate nodes. In contrast to the CBAM (spatial attention block)\cite{woo2018cbam}, since our nodes are not spatially correlated after propagation has forced each node to contain global relations, we do not perform convolution-based squeezing but instead squeeze point-wisely.

\subsection{Conditional Margin Loss: $C$-softmax}
The node passing through the RGM becomes the $L$-dimensional embedding vector through the fully connected layer. In the training phase, the embedding vector goes through the Softmax layer to optimize the loss value through the cross entropy loss function. When testing, the class is predicted by computing the cosine similarity between the embedding vectors of the gallery and the probe images, respectively.
\begin{gather}
	L_{softmax} = -\frac{1}{B}\sum_{i}^{B}log\frac{e^{{\boldsymbol{W}_{k}}^{T}\boldsymbol{x}_{i}+\boldsymbol{b}_{k}}}{\sum_{j}^{M}e^{{\boldsymbol{W}_{j}}^{T}\boldsymbol{x}_{i}+\boldsymbol{b}_{j}}}
	\label{e5}
\end{gather}
\begin{figure}[!t]
	\centering
	\subfloat[CosFace]{\includegraphics[width=0.333\columnwidth]{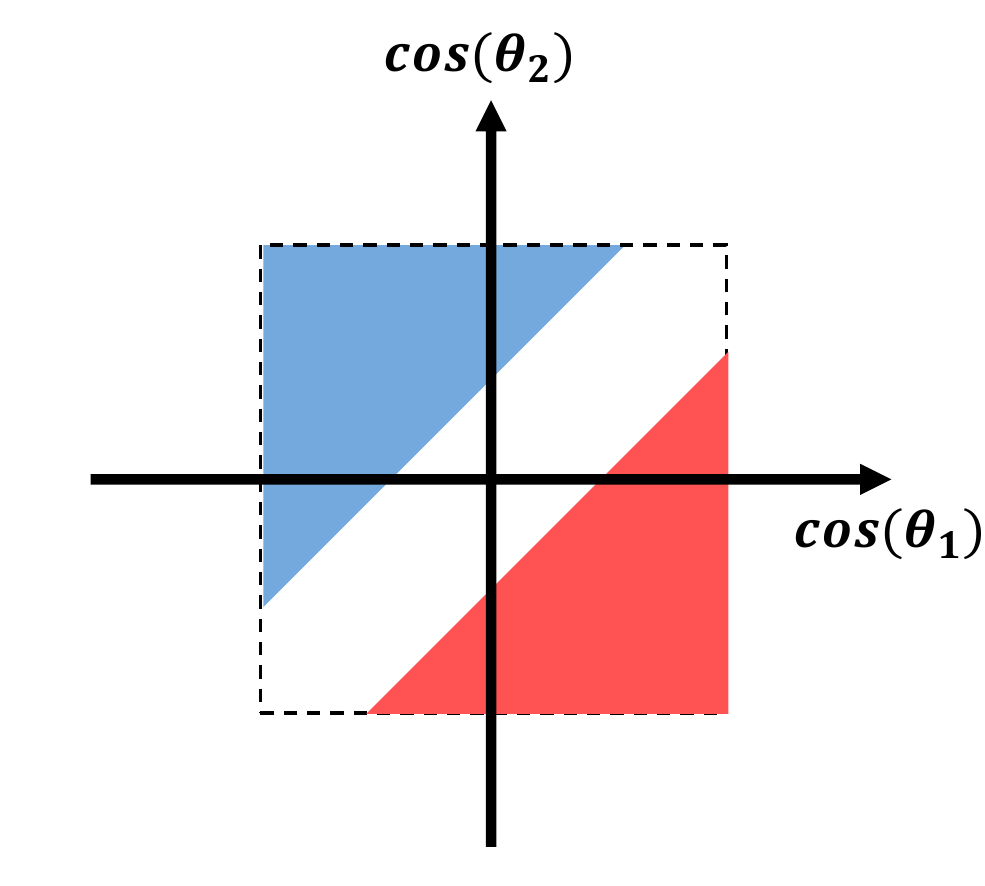}%
		\label{f6_1}}
	\hfil
	\subfloat[ArcFace]{\includegraphics[width=0.333\columnwidth]{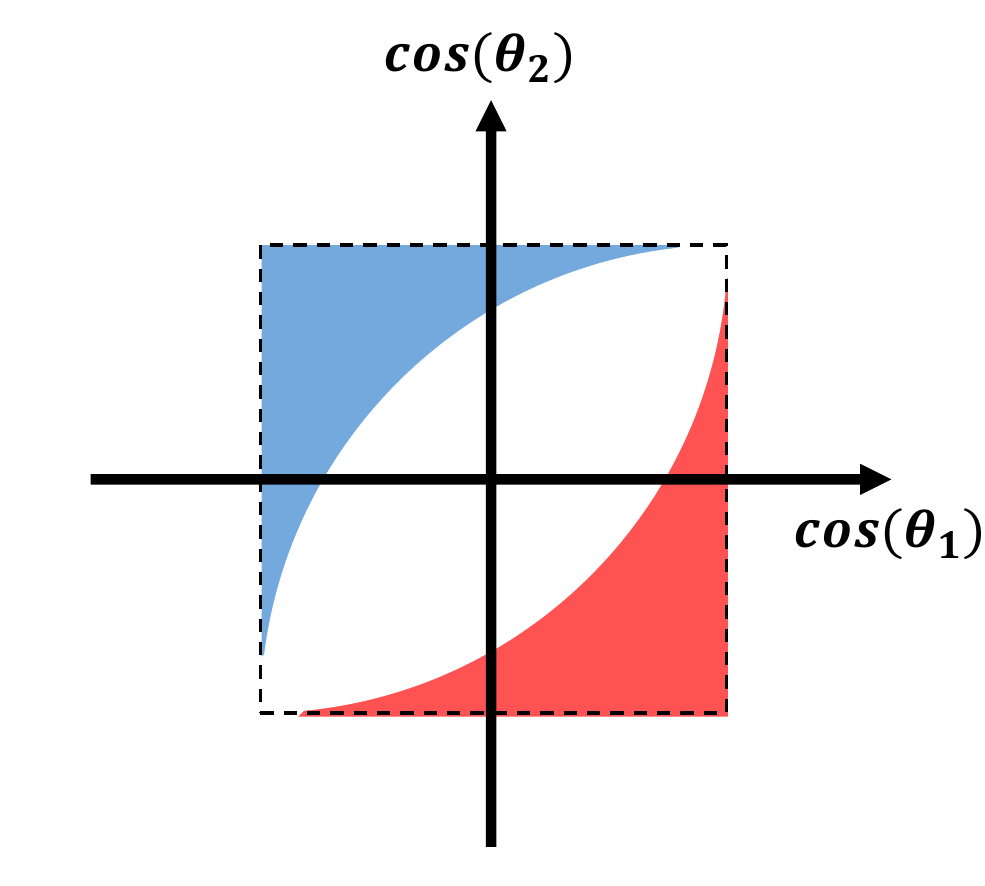}%
		\label{f6_2}}
	\hfil
	\subfloat[$C$-softmax (Ours)]{\includegraphics[width=0.333\columnwidth]{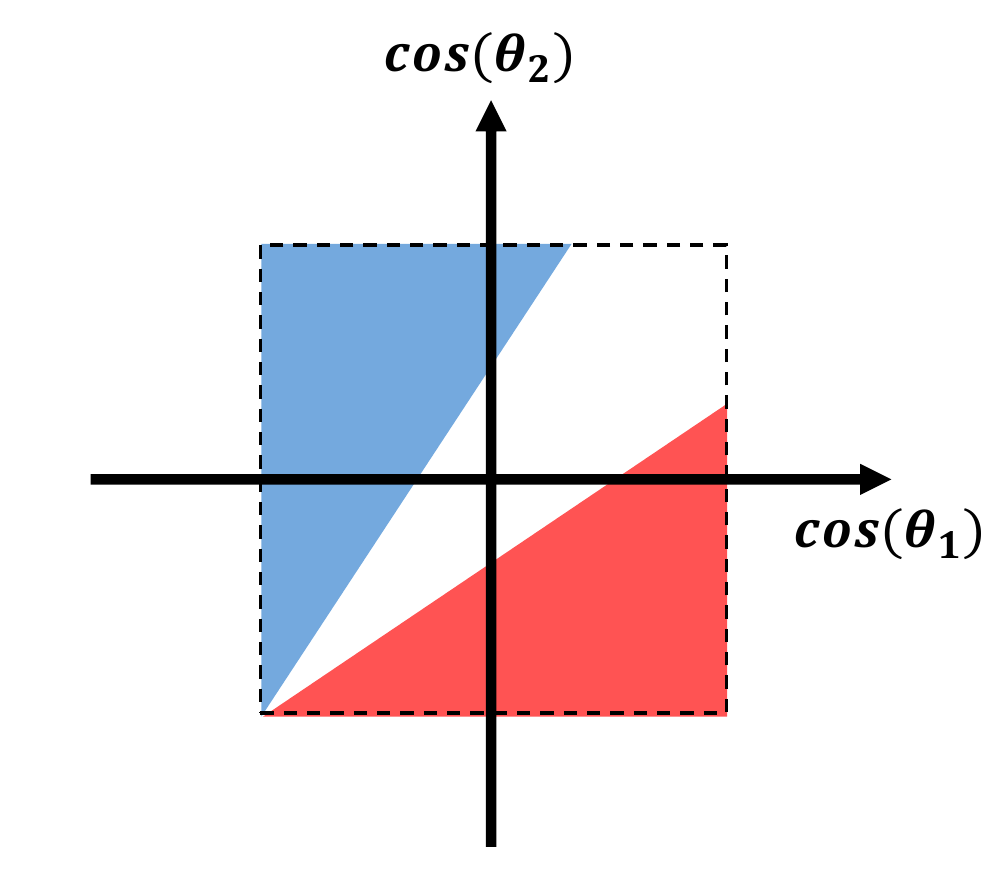}%
		\label{f6_3}}
	\caption{The decision margins of different loss functions within two classes. The X and Y-axes denote the cosine similarities to each class. The red and blue regions correspond to classes 1 and 2, respectively. The white region indicates the margin between classes. In $C$-softmax, when cosine similarities are high in both classes, margin becomes large and margin becomes small when cosine similarities are low.}
	\label{f6}
\vspace{-3mm}
\end{figure}
Equation~\ref{e5} defines a softmax function in which $B$ denotes batch size, $M$ is the number of classes and $x_{i}$ is the embedding vector of the ($k$-th class) training sample. 

In \cite{myeong2019rm}, a triplet loss function with conditional margin is proposed that applies an conditional margin between inter-classes to reduce intra-class discrepancies. This loss function is defined in Equation~\ref{e6}.
\begin{gather}
\left \{ {\boldsymbol{x}_{i}^{a}}, {\boldsymbol{x}_{i}^{p}},{\boldsymbol{x}_{i}^{n}} \right \}\in T \nonumber\\
s_{p} = CS({\boldsymbol{x}_{i}^{a}}, {\boldsymbol{x}_{i}^{p}}) \nonumber\\
s_{n} = CS({\boldsymbol{x}_{i}^{a}}, {\boldsymbol{x}_{i}^{n}}) \nonumber\\
L_{triplet condtional} = \sum_{i}^{N}\left [ \frac{s_{n}+1}{s_{p}+1}-m \right ]_{+} 
\label{e6}
\end{gather}
${\boldsymbol{x}_{i}}^{a}$, ${\boldsymbol{x}_{i}}^{p}$ and ${\boldsymbol{x}_{i}}^{n}$ denotes anchor, positive and negative embedding vector, respectively, where anchor and positive are from same identity while negative sample is from different identity. To reduce domain discrepancy, ${\boldsymbol{x}_{i}}^{a}$ and ${\boldsymbol{x}_{i}}^{p}$ are sampled from different domains. $CS(\cdot)$ indicates cosine similarity and ${S}_{p}$ and ${S}_{n}$ are intra-class and inter-class similarity, respectively. The loss value is calculated from similarity ratios with margin $m$ and this margin considers the distributions of ${S}_{p}$ and ${S}_{n}$.
\begin{gather}
s_{n}<m_{1}s_{p}+m_{2} 
\label{e7}
\end{gather}
This triplet loss function with conditional margin is designed to satisfy Equation~\ref{e7} which takes into account not only the intercept value $m_{2}$ but also the slope $m_{1}$, meaning that every margin is computed conditionally.

Since this loss function utilizes triplet loss, the positive and negative samplings play an important role in learning; therefore online sampling should be done within a mini-batch and semi-hard example learning is required. This increases the training time and makes sampling difficult since the HFR database has only a small number of images and identities. To avoid sampling, we suggest $C$-softmax (conditional-margin softmax) as a loss function, since it applies the margin into the Sofmax layer conditionally according to inter-class similarity. 
\begin{gather}
\boldsymbol{W}^{'}=\frac{\boldsymbol{W}}{\left \| \boldsymbol{W} \right \|}, \boldsymbol{x}^{'}=\frac{\boldsymbol{x}}{\left \| \boldsymbol{x} \right \|} \nonumber\\[5pt]
{\boldsymbol{W}^{'}_{j}}^{T}\boldsymbol{x}^{'}=\frac{\boldsymbol{W}^{T}\boldsymbol{x}}{\left \| \boldsymbol{W} \right \|\left \| \boldsymbol{x} \right \|}=cos\theta _{j} \nonumber\\[5pt]
cos\theta _{j}< m_{1}cos\theta _{i}+m_{2} 
\label{e8}
\end{gather}
\begin{gather}
L_{cond} = -\frac{1}{N}\sum_{i}^{N}log\frac{e^{\alpha({m_{1}cos\theta_{i}+m_{2}})}}{\sum_{j\neq i}^{M}e^{scos\theta_{j}}+ e^{\alpha{(m_{1}cos\theta_{i}+m_{2}})}}
\label{e9}
\end{gather}
First, we normalize the fully connected layer $\boldsymbol{W}$ and embedding vector $\boldsymbol{x}$; normalized vectors are re-scaled to scale $\alpha$, following \cite{ranjan2017l2}. In Equation~\ref{e8}, the product of these two vectors gives the angle between the two vectors, which defines cosine similarity. Therefore, the conditional margin in Equation~\ref{e7} can be written as Equation~\ref{e8} and it transforms as Equation~\ref{e9}. Here $L_{cond}$ is $C$-softmax loss function and $m_{1}$ and $m_{2}$ indicate the slope and intercept values respectively, as in Equation~\ref{e7}. 
\begin{figure}[t]
	\centering
	\subfloat[CosFace]{\includegraphics[width=0.325\columnwidth]{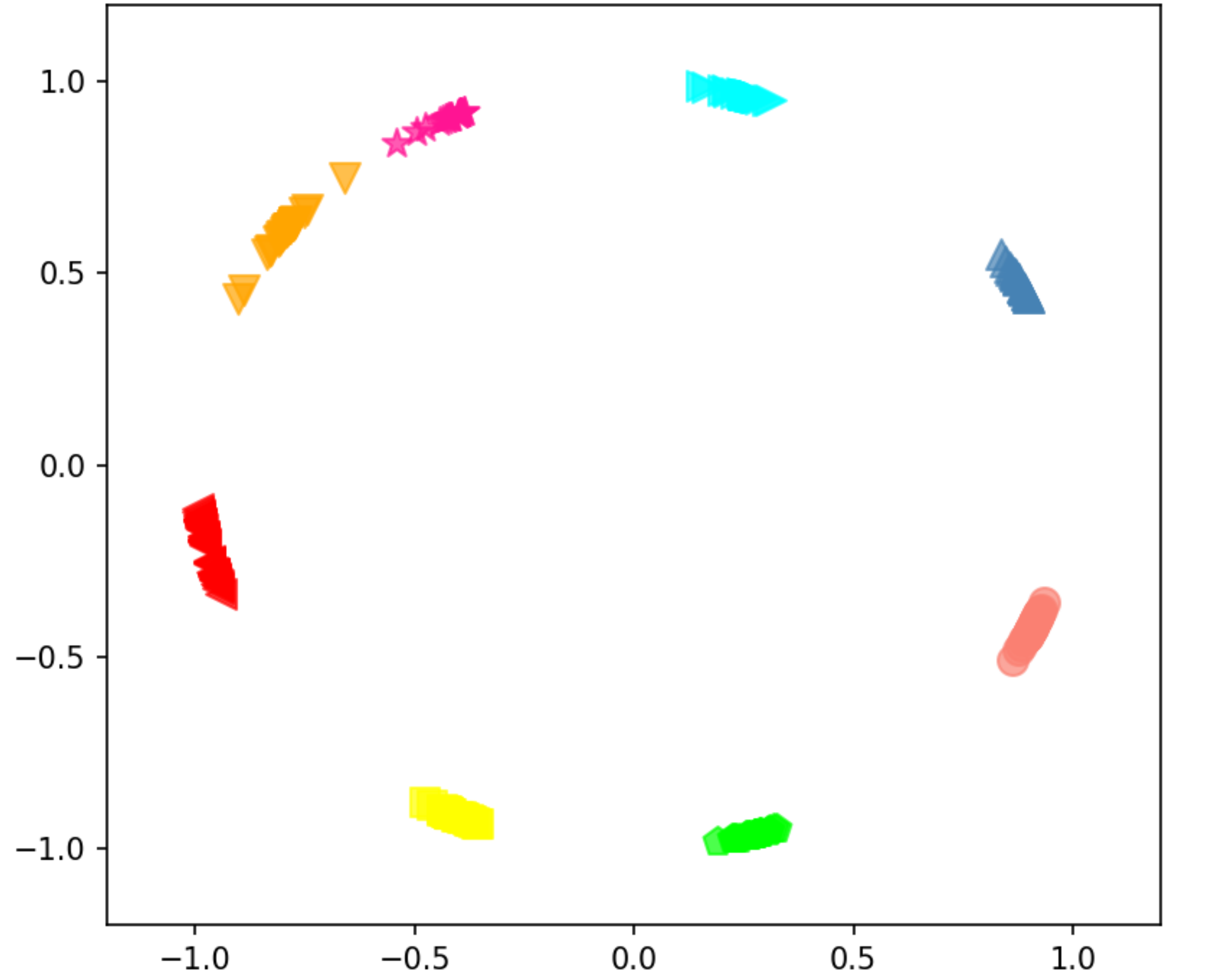}%
		\label{f_cos}}
	\hfil
	\subfloat[ArcFace]{\includegraphics[width=0.325\columnwidth]{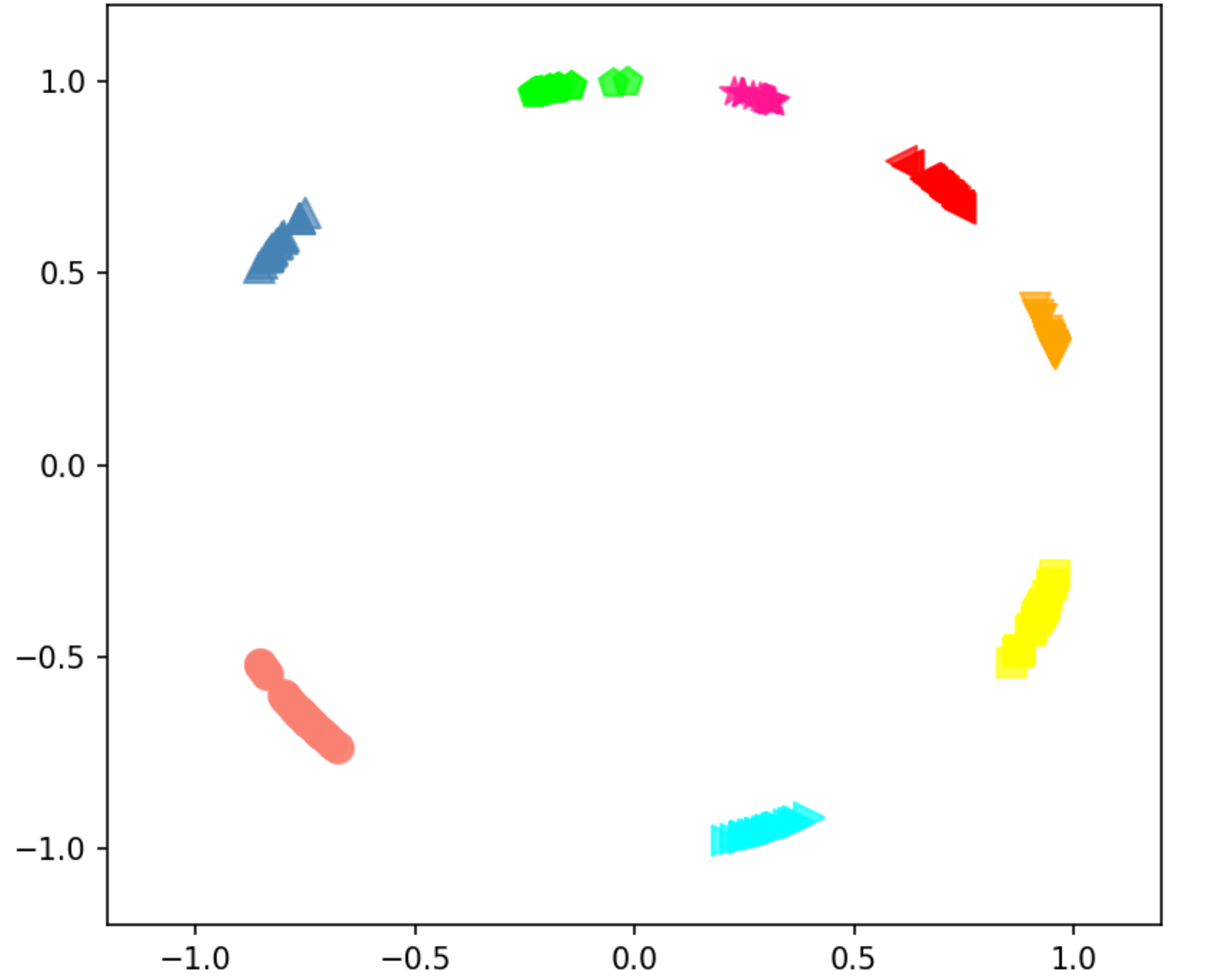}%
		\label{f_arc}}
	\hfil
	\subfloat[$C$-softmax]{\includegraphics[width=0.35\columnwidth]{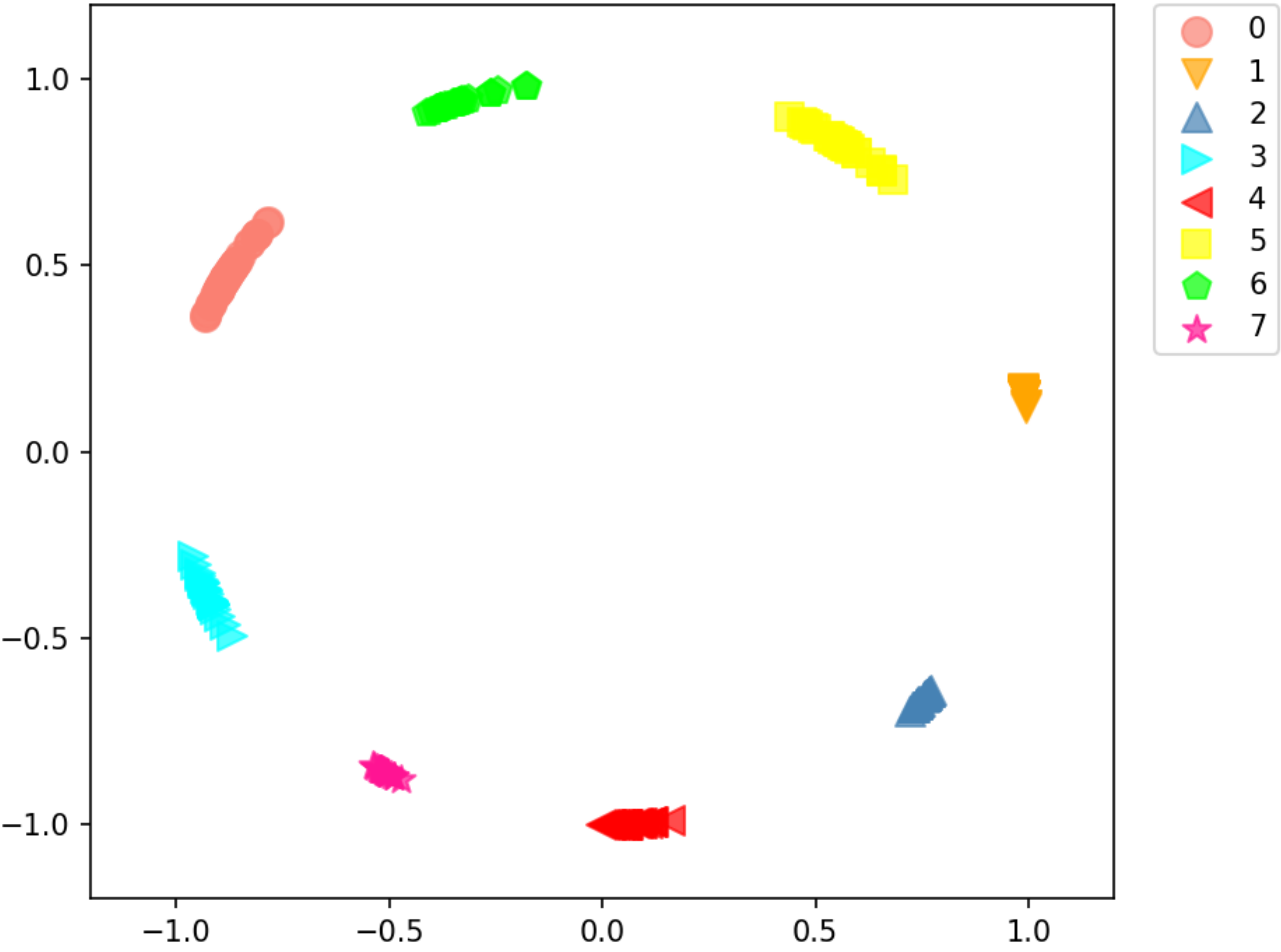}%
		\label{f_cond}}
	\caption{A toy experimental result of different angular loss functions. 2D feature network is trained on eight subjects from CASIA NIR-VIS 2.0 database. 2D features of training samples are projected in angular space and each color represents a subject.}
	\label{f_toy}
	\vspace{-3mm}
\end{figure}

Figure~\ref{f6} shows the margin according to each cosine similarity in two classes. Compared to CosFace\cite{wang2018cosface} and ArcFace\cite{deng2019arcface}, our proposed conditional margin is determined conditionally by considering the similarity between classes. When this similarity is small, sufficient inter-class space is guaranteed, so that we do not need to set a high margin. Conversely, when the similarity between classes is high, we have a hard class example, so the margin should be increased to give a stricter criterion. In this way, the margin can be conditionally determined according to the similarity value, which gives a hard sampling effect by concentrating on the hard sample (see Figure~\ref{f6}(c)). When we decrease $m_{1}$, the margin at the large similarity region increases since the slope is gentle; when we increase $m_{2}$, the margin at the small similarity region increases. To prevent a negative margin, we use a constraint such as $m_{1}-m_{2} \geq 1$. By contrast, CosFace\cite{wang2018cosface} gives a constant margin for cosine similarity using $cos\theta_{j}< cos\theta_{i}+m$, and ArcFace\cite{deng2019arcface} uses a constant margin for class angular domain with $cos\theta_{j}< cos(\theta_{i}+m)$. When we convert the ArcFace class angular domain $\theta$ to $cos\theta$, as shown in Figure~\ref{f6}(b), the margin varies depending on the similarity, but a large margin occurs only when the similarity is near the midpoint. In contrast, our margins are given conditionally so that heterogeneous data with large intra-class discrepancy can be more efficiently trained during common space learning. 

Figure~\ref{f_toy} is a toy experiment result of angular losses. We train a 2D feature embedding network with eight subjects of face images and visualized 2D features of training images, following \cite{wang2018cosface, deng2019arcface}. The training samples are from CASIA NIR-VIS 2.0 (around 30 NIR and VIS images per class). $C$-softmax projects samples into a compact intra-class and a large inter-class by providing different margins for inter-class similarity, but in case of other losses, a large inter-class is not guaranteed since they provide a fixed margin value. This is because two classes of high similarity are projected closer than two classes of low similarity. Through the margin based on the similarity between two classes (small margin for small similarity and large margin for large similarity), each class is finally mapped at the almost identical intervals. 

\section{Experimental Results} \label{Experimental results}
In this section, we experiment proposed method on five HFR databases, namely CAISA NIR-VIS 2.0\cite{li2013casia}, IIIT-D Sketch\cite{bhatt2012memetic}, BUAA-VisNir\cite{huang2012buaa}, Oulu-CASIA NIR-VIS\cite{zhao2011facial}, and TUFTS\cite{panetta2018comprehensive}. For each database, we perform ablation studies and comparison with other state-of-the-art methods. Also, we compare the RGM with other attentional modules and $C$-softmax loss function with other angular margin losses. Finally, we analyze and discuss the visualization of the extracted relational information of RGM and NAU.

Our three backbones are LightCNN-9, LightCNN-29\cite{wu2018light}, and ResNet18\cite{he2016deep}, consisting of 9, 29 and 18 convolutional layers respectively. These three baseline networks are pre-trained on the MS-Celeb-1M, a large-scale visual face database. To fine-tune the proposed modules, the pre-trained feature extractor is frozen and only the HFR database, comprising non-VIS and VIS faces, is used for training data. For the fair comparison, since learnable parameters are added by RGM and NAU, we add extra convolutional layers after the backbone in which a number of parameters are similar to RGM and NAU following \cite{deng2019residual}. Two extra convolutional layers $(128, 128, 1\times1)$ are added at LightCNN-9 and LightCNN-29, and one extra convolutional layer $(512, 512, 1\times1)$ is added at ResNet18. The numbers in parentheses indicate input channel, output channel, and filter size, respectively. We use 128 (or 64 for small database) batches and learning rate starts at 0.001 (or 0.01 for the IIIT-D Sketch database); to avoid over-fitting, the dropout\cite{srivastava2014dropout} rate is set to 0.7 at the fully connected layer. For an input, we crop each image to $144\times 144$ size and randomly crop to $128 \times 128$ for LightCNN and $112 \times 112$ for ResNet. The RGM is plugged after the last convolutional layer and use 64 ($8 \times 8$) node vectors. In the NAU, the channel reduction ratio is 2; in $C$-softmax loss function $m_{1}=0.7$, $m_{2}=-0.3$, and the normalized scale value $s$ = 24 are used.

\begin{figure}[t]
	\centering
	\includegraphics[width=0.85\linewidth]{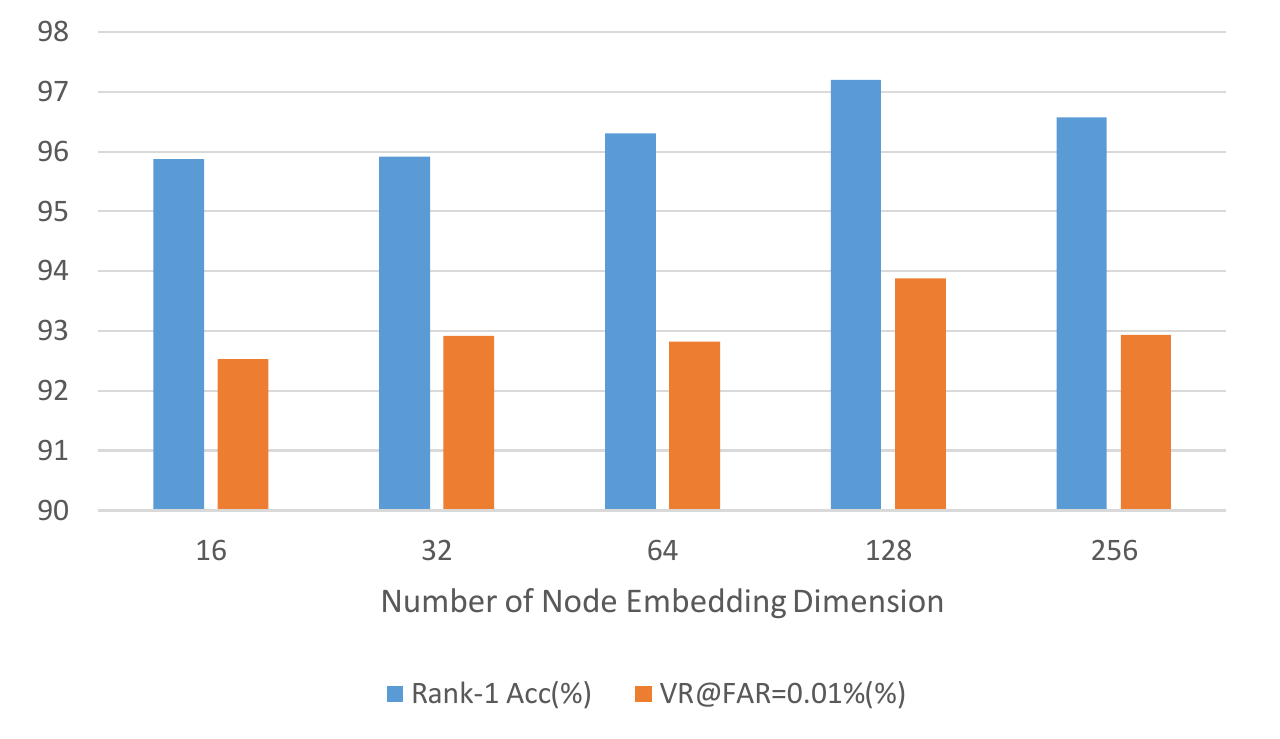}
	\caption{The rank-1 accuracy rate in CASIA NIR-VIS 2.0 database according to the node embedding vector dimension.}
	\label{f7}
\vspace{-3mm}
\end{figure}

\subsection{CASIA NIR-VIS 2.0}
\subsubsection{Database}
The CASIA NIR-VIS 2.0 database is one of the largest HFR databases, and is composed of NIR and VIS face images. It contains 725 subjects, imaged by VIS and NIR cameras in four recording sessions. 
We followed view 2 protocol, which training subjects and the corresponding testing sets are non-overlapped and the numbers of subjects are virtually identical. For evaluation, the gallery set comprises one VIS image per subject while the probe set contains several NIR images per subject. The prediction score is computed by similarity matrix over the whole gallery set and the identification accuracy and verification rate recorded. 

\subsubsection{Ablation Studies} \label{ablation}
We first experiment with different numbers of the node vector dimension $d$ to find the appropriate dimension for HFR. The experiment is conducted on LightCNN-9, in which there are 128 channels in the last convolutional layer. Figure~\ref{f7} shows the results of training with the RGM node vector dimension $ d = 16, 32, 64, 128$ and $256 $. The identification accuracy and verification rate show better performance as the dimension increases and then drop off when it becames too large. We use a dimension of 128 in LightCNN and 256 in ResNet18, whose channel size at the last convolutional layer is $512$.

\begin{table}[t]
	\centering
	\caption{Ablation studies of the proposed method on the CASIA NIR-VIS 2.0 database}
	\resizebox{\linewidth}{!}{
	\renewcommand{\arraystretch}{1.0}
	\footnotesize
	\begin{tabular}{cccccc}
		\hline
		\multicolumn{2}{c}{\multirow{2}[1]{*}{Models}} & \multicolumn{1}{p{4.05em}}{Rank-1 \newline{}Acc(\%)} & \multicolumn{1}{p{4.05em}}{VR@FAR\newline{}=1\%(\%)} & \multicolumn{1}{p{4.05em}}{VR@FAR\newline{}=0.1\%(\%)} & \multicolumn{1}{p{4.05em}}{VR@FAR\newline{}=0.01\%(\%)} \\
		\midrule
		\midrule
		\multirow{5}[4]{*}{\begin{sideways}LightCNN-9\end{sideways}} & fine-tuned & 93.21 & 98.01 & 93.41 & 90.15 \\
		& +extra conv & 96.91 & 98.83 & 95.48 & 93.68 \\
		\cmidrule{2-6} & RGM   & 96.7  & 98.86 & 95.66 & 93.43 \\
		& +NAU  & 97.2  & 98.76 & 95.79 & 93.9 \\
		& +$C$-softmax & \textbf{98.03} & \textbf{99.15} & \textbf{96.76} & \textbf{95.23} \\
		\midrule
		\multirow{5}[4]{*}{\begin{sideways}ResNet18\end{sideways}} & fine-tuned & 88.87 & 91.24 & 79.97 & 74.99 \\
		& +extra conv & 94.73 & 97.76 & 94.15 & 92.14 \\
		\cmidrule{2-6} & RGM   & 96.33 & 98.59 & 96.39 & 94.95 \\
		& +NAU  & 96.67 & 98.71 & 96.47 & 95.07 \\
		& +$C$-softmax & \textbf{97.44} & \textbf{98.79} & \textbf{96.71} & \textbf{95.43} \\
		\midrule
		\multirow{5}[3]{*}{\begin{sideways}LightCNN-29\end{sideways}} & fine-tuned & 97.65 & 99.34 & 97.79 & 96.84 \\
		& +extra conv & 98.85 & 99.47 & 98.33 & 97.72 \\
		\cmidrule{2-6} & RGM   & 98.98 & 99.5  & 95.65 & 98.05 \\
		& +NAU  & 99.06 & \textbf{99.94} & \textbf{99.5} & 98.11 \\
		& +$C$-softmax & \textbf{99.3} & 99.51 & 99.02 & \textbf{98.86} \\[0.1cm]
		\hline
	\end{tabular}}
	\label{t1}%
\vspace{-3mm}
\end{table}
%
Table~\ref{t1} shows ablation studies in three baseline networks on the CASIA NIR-VIS 2.0 database. In the table, the fine-tuned indicates the results of training only the fully connected layers while freezing the pre-trained feature extractor. For each network we attach our proposed RGM module, then experiment with NAU, and finally show the results of training with $C$-softmax. The ResNet18 fine-tuned is 88.37\% in rank-1 accuracy. When the RGM extracts domain-invariant features focused on relational information, the performance improves to 96.33\%. When the network trains with NAU and $C$-softmax, it shows additional performance improvement of 0.34\% and 0.77\% respectively which is higher than baseline with extra convolutional layer. Similarly, performance on LightCNNs are improved by 4.82\% and 1.65\% over fine-tuned accuracy. 

\subsubsection{Comparison with Other Methods}
In Table~\ref{t2}, we compare our method (LigthCNN-29 backbone) with other deep learning-based HFR methods, namely HFR-CNN\cite{saxena2016heterogeneous}, TRIVLET\cite{liu2016transferring}, ADFL\cite{song2018adversarial}, CDL\cite{wu2018coupled}, WCNN\cite{he2018wasserstein}, DSU\cite{de2018heterogeneous}, RCN\cite{deng2019residual} and RM\cite{myeong2019rm}. All comparison methods use backbone pre-trained on MS-Celeb-1M or CASIA WebFace. Since RCN\cite{deng2019residual} performance in original paper used a backbone pre-trained on five large-scale databases, we reproduced RCN with LightCNN-29 pre-trained on MS-Celeb-1M as a backbone for fair comparison. The RM method, which extracts features by pair-wise relation embedding, performs better than other deep learning methods. Our method shows 0.38\% performance improvement over RM and also yields results comparable with other domain-invariant based methods.

\subsubsection{Comparison with Attentional Modules}  \label{relational method}
In Table~\ref{t3}, we compare RGM with the attentional modules depicted earlier in Figure~\ref{f3}. We train under the same conditions, with the last feature map of LightCNN-9 and LigthCNN-29 passed through each module with cross entropy loss.
The RM and graph-structured modules: GloRe \cite{chen2019graph} and RGM show higher performance than other methods, among which RGM shows 1.02\% and 0.24\% higher performance over second best performance in each backbone.
\begin{table}[t]
	\begin{center}
	\caption{Comparison with Other Methods on the 10-fold CASIA NIR-VIS 2.0 Database}
	\begin{tabular}{ccc}
		\hline
		\multicolumn{1}{c}{Methods} & Rank-1 \newline{}Acc(\%) & VR@FAR\newline{}=0.1\%(\%) \\
		\hline 
		\hline
		HFR-CNN\cite{saxena2016heterogeneous} & 85.9 $\pm $ 0.9 & \multicolumn{1}{c}{78.0} \\
		TRIVLET\cite{liu2016transferring} & 95.7 $\pm $ 0.5 & 91.0 $\pm $ 1.3 \\
		ADFL\cite{song2018adversarial} & 98.2 $\pm $ 0.3 & 97.2 $\pm $ 0.3 \\
		CDL\cite{wu2018coupled} & 98.6 $\pm $ 0.2 & 98.3 $\pm $ 0.1 \\
		WCNN\cite{he2018wasserstein} & 98.7 $\pm $ 0.3 & 98.4 $\pm $ 0.4 \\
		DSU\cite{de2018heterogeneous} & 96.3 $\pm $ 0.4 & 98.4 $\pm $ 0.12 \\
		RCN\cite{deng2019residual} & 98.48  $\pm $ 0.5  & 97.77  $\pm $ 0.4 \\
		\textbf{RM\cite{myeong2019rm}} & \multicolumn{1}{c}{98.92 $\pm $ 0.16} & \multicolumn{1}{c}{98.72 $\pm $ 0.2} \\
		\textbf{Ours} & \textbf{99.3 $\pm $0.1} & \textbf{98.9 $\pm $ 0.12} \\
		\hline
	\end{tabular}
	\label{t2}%
	\end{center}
\vspace{-3mm}
\end{table}%

\begin{table}[t]
	\centering
	\caption{Comparison with attentional modules on the CASIA NIR-VIS 2.0 database}
	\resizebox{\linewidth}{!}{
		\begin{tabular}{ccccc}
			\hline
			\multicolumn{1}{c}{Backbone} & \multicolumn{2}{c}{LightCNN-9} & \multicolumn{2}{|c}{LightCNN-29} \\
			\hline
			\multicolumn{1}{c}{\multirow{2}[1]{*}{Modules}} & \multicolumn{1}{p{4.05em}}{Rank-1 \newline{}Acc(\%)} & \multicolumn{1}{p{4.05em}}{VR@FAR\newline{}=0.1\%(\%)} & \multicolumn{1}{|p{4.05em}}{Rank-1 \newline{}Acc(\%)} & \multicolumn{1}{p{4.05em}}{VR@FAR\newline{}=0.1\%(\%)} \\
			\hline
			\hline
			finetune & 93.21 & 93.41 & \multicolumn{1}{|c}{97.65} & 97.79 \\
			B-CNN\cite{chowdhury2016one} & 81.67 & 82.17 & \multicolumn{1}{|c}{92.04} & 91.05 \\
			Non-Local\cite{wang2018non} & 92.51 & 92.11  & \multicolumn{1}{|c}{98.98} & 98.37 \\
			DoubleAttention\cite{chen20182} & 70.48 & 68.29 & \multicolumn{1}{|c}{88.24} & 88.11 \\
			GloRe\cite{chen2019graph} & 96.18 & 94.68 & \multicolumn{1}{|c}{98.82} & 98.33 \\
			\textbf{RM\cite{myeong2019rm}} & 94.73 & 94.31 & \multicolumn{1}{|c}{98.12} & 97.68 \\
			\textbf{RGM(Ours)} & \textbf{97.2} & \textbf{95.79} & \multicolumn{1}{|c}{\textbf{99.06}} & \textbf{99.5} \\
			\hline
	\end{tabular}}
	\label{t3}%
	\vspace{-3mm}
\end{table}

\begin{table*}[t]
	\centering
	\caption{Ablation studies of the proposed method on the IIIT-D Sketch, BUAA Vis-Nir and Oulu-CASIA NIR-VIS database}
	\resizebox{\textwidth}{!}{
		\begin{tabular}{cccccccccc}
			\hline
			\multicolumn{2}{c}{\multirow{2}[1]{*}{Models}} & \multicolumn{3}{c}{IIIT-D Sketch} & \multicolumn{2}{|c}{BUAA-VisNir}&\multicolumn{3}{|c}{Oulu-CASIA NIR-VIS} \\
			\multicolumn{2}{c}{} & \multicolumn{1}{c}{Rank-1 Acc(\%)} & \multicolumn{1}{c}{VR@FAR=1\%(\%)} & \multicolumn{1}{c}{VR@FAR=0.1\%(\%)} & \multicolumn{1}{|c}{Rank-1 Acc(\%)} & \multicolumn{1}{c}{VR@FAR=1\%(\%)} & \multicolumn{1}{|c}{Rank-1 Acc(\%)} & VR@FAR=1\%(\%) & VR@FAR=0.1\%(\%)\\
			\hline
			\hline
			\multirow{5}[4]{*}{LightCNN-9} & fine-tuned & 78.72 & 92.84 & 89.75 & \multicolumn{1}{|c}{94.78} & 88.22 &\multicolumn{1}{|c}{96.35}&97.81&95.21\\
			& +extra conv & 85.11 & 97.02 & 94.04 & \multicolumn{1}{|c}{93.33} & 94.11 &\multicolumn{1}{|c}{98.54}&96.77&96.04\\
			\cline{2-10}  & RGM   & 88.08 & 99.78 & 94.47 & \multicolumn{1}{|c}{92.67} & 87.33 &\multicolumn{1}{|c}{98.44}&96.88&93.65\\
			& +NAU  & \textbf{88.94} & \textbf{97.87} & \textbf{95.74} & \multicolumn{1}{|c}{95.11} & 88.44&\multicolumn{1}{|c}{99.27}&98.44&96.77 \\
			& +$C$-softmax & 88.51 & 96.17 & 94.47 & \multicolumn{1}{|c}{\textbf{97.56}} & \textbf{98.1} &\multicolumn{1}{|c}{\textbf{99.27}}&\textbf{99.69}&\textbf{98.96}\\
			\hline
			\multirow{5}[4]{*}{ResNet18} & fine-tuned & 70.21 & 86.81 & 82.55 & \multicolumn{1}{|c}{97.67} & 97.33 &\multicolumn{1}{|c}{99.17}&96.77&95.83\\
			& +extra conv & 83.83 & 95.32 & 94.89 & \multicolumn{1}{|c}{95.89} & 95.44 &\multicolumn{1}{|c}{100.0}&98.75&97.71\\
			\cline{2-10} & RGM   & 85.11 & 95.41 & 94.89 & \multicolumn{1}{|c}{\textbf{99.22}} & \textbf{98.22} &\multicolumn{1}{|c}{99.9}&97.92&94.9\\
			& +NAU  & 85.11 & 95.74 & 94.47 & \multicolumn{1}{|c}{98.89} & 97.11 &\multicolumn{1}{|c}{100.0}&\textbf{99.17}&98.96\\
			& +$C$-softmax & \textbf{85.96} & \textbf{95.74} & \textbf{95.32} & \multicolumn{1}{|c}{99}    & 97.22 &\multicolumn{1}{|c}{\textbf{100.0}}&98.96&\textbf{99.17}\\
			\hline
			\multirow{5}[3]{*}{LightCNN-29} & fine-tuned & 62.98 & 84.68 & 81.7  & \multicolumn{1}{|c}{97.44} & 98.89&\multicolumn{1}{|c}{99.27} &\textbf{99.69}&98.96\\
			& +extra conv & 74.04 & 91.49 & 90.64 & \multicolumn{1}{|c}{99.11} & 99.44&\multicolumn{1}{|c}{100.0}&99.06&98.12\\
			\cline{2-10} & RGM   & 74.5  & 92.77 & 91.06 & \multicolumn{1}{|c}{99.56} & 99.22 &\multicolumn{1}{|c}{100.0}&98.44&96.88\\
			& +NAU  & 78.72 & \textbf{94.47} & \textbf{92.34} & \multicolumn{1}{|c}{99.56} & 99.11&\multicolumn{1}{|c}{100.0}&98.44&96.88 \\
			& +$C$-softmax & \textbf{79.15} & 94.04 & 91.49 & \multicolumn{1}{|c}{\textbf{99.67}} & \textbf{99.22} &\multicolumn{1}{|c}{\textbf{100.0}}&99.17&\textbf{98.96}\\
			\hline
	\end{tabular}}
	\label{t5}%
\vspace{-1mm}
\end{table*}%

\begin{table}[ht]
	\begin{center}
	\caption{Comparison with other methods on the IIIT-D Sketch database}
	\renewcommand{\arraystretch}{1}
	\begin{tabular}{ccc}
		\hline
		Model & \multicolumn{1}{c}{Rank-1 \newline{}Acc(\%)} & \multicolumn{1}{c}{VR@FAR\newline{}=1\%(\%)} \\
		\hline
		\hline
		SIFT\cite{bhatt2012memetic} & 76.28 & - \\
		\multicolumn{1}{l}{MCWLD\cite{bhatt2012memetically}} & \multicolumn{1}{c}{84.24} & - \\
		VGG\cite{parkhi2015deep} & 80.89 & 72.08 \\
		CenterLoss\cite{wen2016discriminative} & 84.07 & 76.2 \\
		CDL\cite{wu2018coupled} & 85.35 & 82.52 \\
		RCN\cite{deng2019residual} & 63.83 & 90.12 \\
		\textbf{RM\cite{myeong2019rm}} & 77.45 & 91.34 \\
		\textbf{Ours}  &\textbf{88.94} & \textbf{97.87} \\
		\hline
	\end{tabular}
	\end{center}
	\label{t4}%
\vspace{-2mm}
\end{table}

\begin{table}[ht]
	\centering
	\caption{Comparison with attentional modules on the IIIT-D Sketch database}
	\resizebox{\linewidth}{!}{
	\begin{tabular}{ccccc}
		\hline
		\multicolumn{1}{c}{Backbone} & \multicolumn{2}{c}{LightCNN-9} & \multicolumn{2}{|c}{LightCNN-29} \\
		\hline
		\multicolumn{1}{c}{\multirow{2}[1]{*}{Modules}} & \multicolumn{1}{p{4.05em}}{Rank-1 \newline{}Acc(\%)} & \multicolumn{1}{p{4.05em}}{VR@FAR\newline{}=1\%(\%)} & \multicolumn{1}{|p{4.05em}}{Rank-1 \newline{}Acc(\%)} & \multicolumn{1}{p{4.05em}}{VR@FAR\newline{}=1\%(\%)} \\
		\hline
		\hline
		finetune & 78.72 & 92.84 & \multicolumn{1}{|c}{62.98} & \textbf{94.68}\\
		B-CNN\cite{chowdhury2016one} & 50.21 & 70.64 & \multicolumn{1}{|c}{26.81} & 44.26 \\
		Non-Local\cite{wang2018non} & 80    & 94.47 & \multicolumn{1}{|c}{70.64} & 91.49 \\
		DoubleAttention\cite{chen20182} & 44.31 & 67.98 & \multicolumn{1}{|c}{27.66} & 48.09 \\
		GloRe\cite{chen2019graph} & 79.15 & 94.89 & \multicolumn{1}{|c}{74.04} &91.49 \\
		\textbf{RM\cite{myeong2019rm}} & 77.45    &     92.34  &  \multicolumn{1}{|c}{65.11} & 88.94\\
		\textbf{RGM(Ours)} & \textbf{88.94} & \textbf{97.87} & \multicolumn{1}{|c}{\textbf{78.72}} & 94.47 \\
		\hline
	\end{tabular}}
	\label{t6}%
\vspace{-3mm}
\end{table}%

\subsection{IIIT-D Sketch}
\subsubsection{Database}
The IIIT-D Sketch database is designed for the sketch-to-photo face recognition task. We use the Viewed Sketch Database which comprises 238 subjects. Each subject has one image pair, a sketch and a VIS photo face image. 
Since there are only a small number of images for training, we train on CUHK Face Sketch FERET Database (CUFSF)\cite{wang2008face} and evaluate on IIIT-D Sketch database, following the same protocol as in \cite{wu2018coupled}. The CUFSF database includes 1,194 subjects from the FERET database \cite{phillips2000feret}, with a single sketch and photo image pair per subject. For testing, we use VIS photo images as the gallery set and sketch images as the probe set. 

\subsubsection{Ablation Studies}
In IIIT-D database, where the domain discrepancy is large and the data is insufficient (only one pair image for each identity), the deeper the layer of the backbone, the lower the performance due to the overfitting problem \cite{deng2019mutual}.
As with the results on CASIA NIR-VIS 2.0 database, in Table~\ref{t5}, our approach improves further with the addition of RGM, NAU and $C$-softmax loss. However, when LightCNN-9 is the baseline, training with the original softmax performs 0.43\% better than with the $C$-softmax. This is because the number of CUFSF and IIIT-D images is smaller than for the CASIA NIR-VIS database, so it is difficult to learn sufficiently with $C$-softmax loss and the margin values $m_{1}$ and $m_{2}$ need to be adjusted. 

\subsubsection{Comparison with Other Methods}
As we described in Table~\ref{t4}, SIFT\cite{bhatt2012memetic}, MCWLD\cite{bhatt2012memetically}, VGG\cite{parkhi2015deep}, CenterLoss\cite{wen2016discriminative}, CDL\cite{wu2018coupled} and RCN\cite{deng2019residual} are compared with our approach and all of these methods are use backbone pre-trained on MS-Celeb-1M (We reproduce RCN with LightCNN-9 pre-trained on MS-Celeb-1M for fair comparison). In particular, the sketch HFR database comprises artiest's pictures, rather than the photos, making training based on deep learning difficult. Nevertheless, our method shows a rank-1 accuracy of 88.94\%, the leading performance among deep learning and hand-crafted methods with same pre-trained on MS-Celeb-1M database condition.

\subsubsection{Comparison with Attentional Modules}
We also apply the attention method and the graph method on LightCNN-9 and LightCNN-29 to the sketch-to-photo HFR task (Table~\ref{t6}). As with the NIR database, the B-CNN and DoubleAttention module show low performance making it difficult to reduce the sketch domain discrepancy via the self-attention method by simply multiplying feature vectors. 
In the Sketch database, which has a large domain difference and a small number of images, the RGM shows higher performance than second best performance by a larger amount of 9.79\% and 4.68\% on each backbone. The RGM, which extracts relational information with small parameters, prevents overfitting and outperforms even with a small database.

\begin{figure*}[!t]
	\centering
	\subfloat[]{\includegraphics[width=0.22\textwidth]{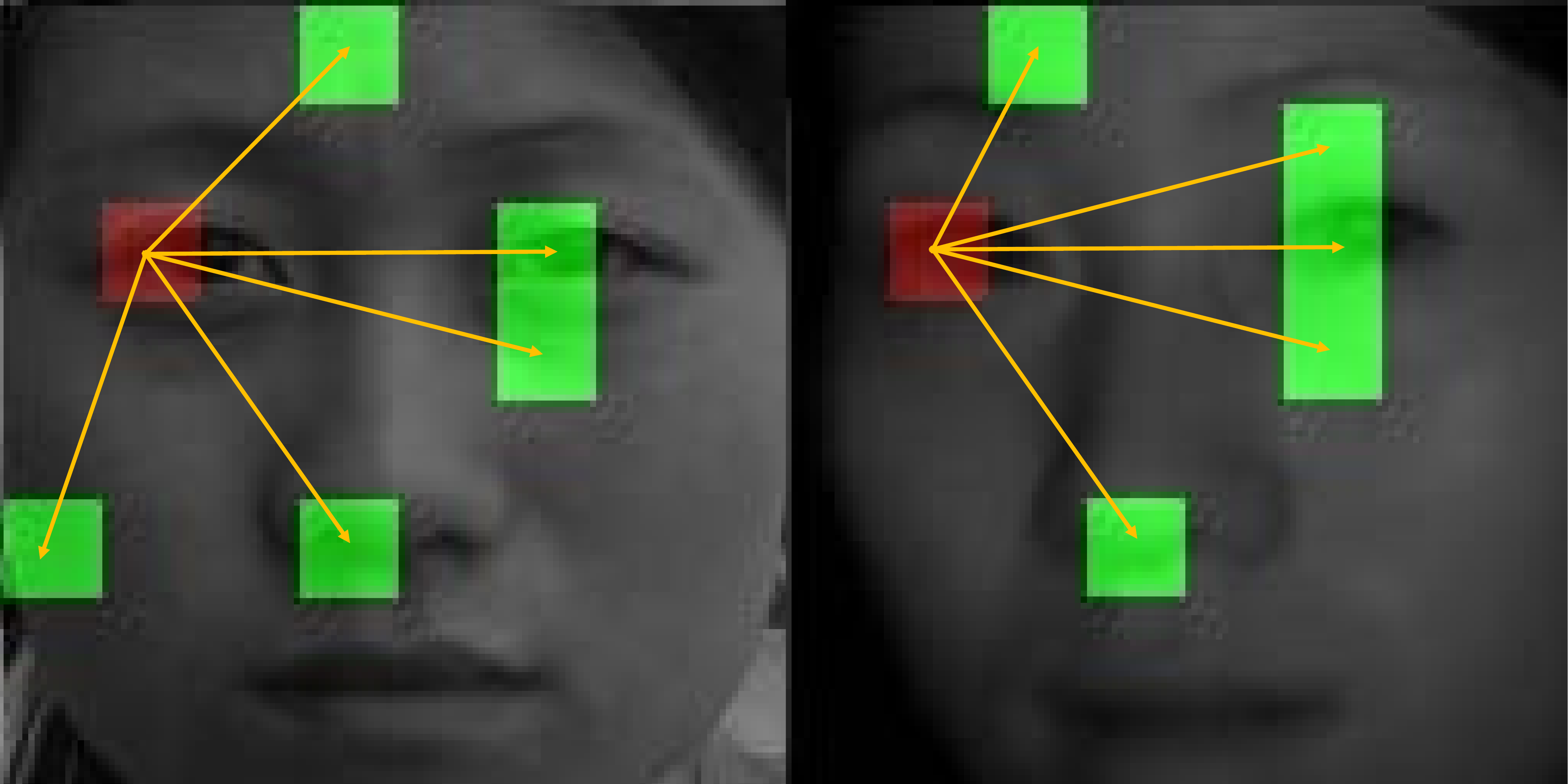}%
		\label{f8_re_a}}
	\hfil
	\subfloat[]{\includegraphics[width=0.22\textwidth]{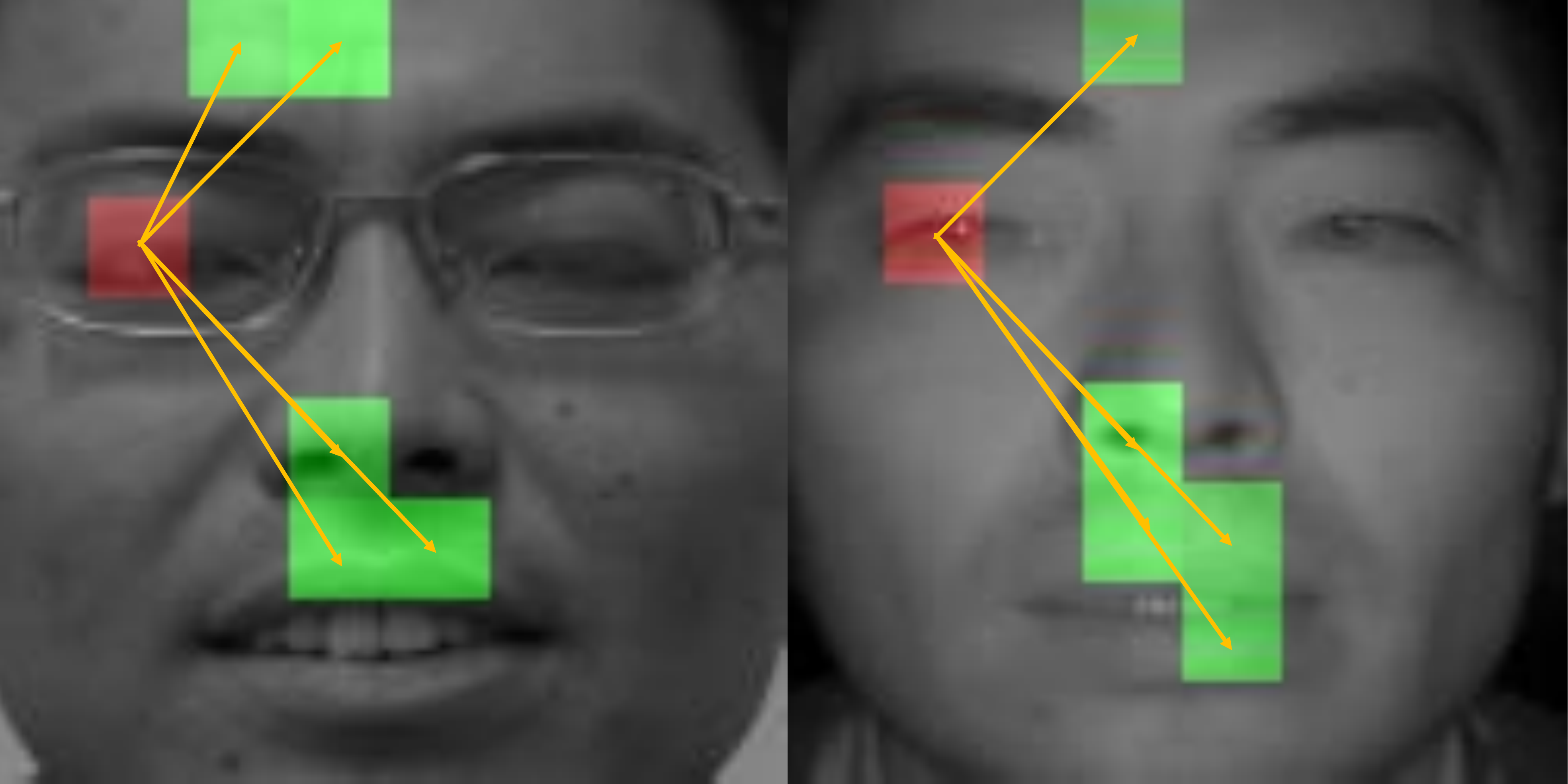}%
		\label{f8_re_b}}
	\hfil
	\subfloat[]{\includegraphics[width=0.22\textwidth]{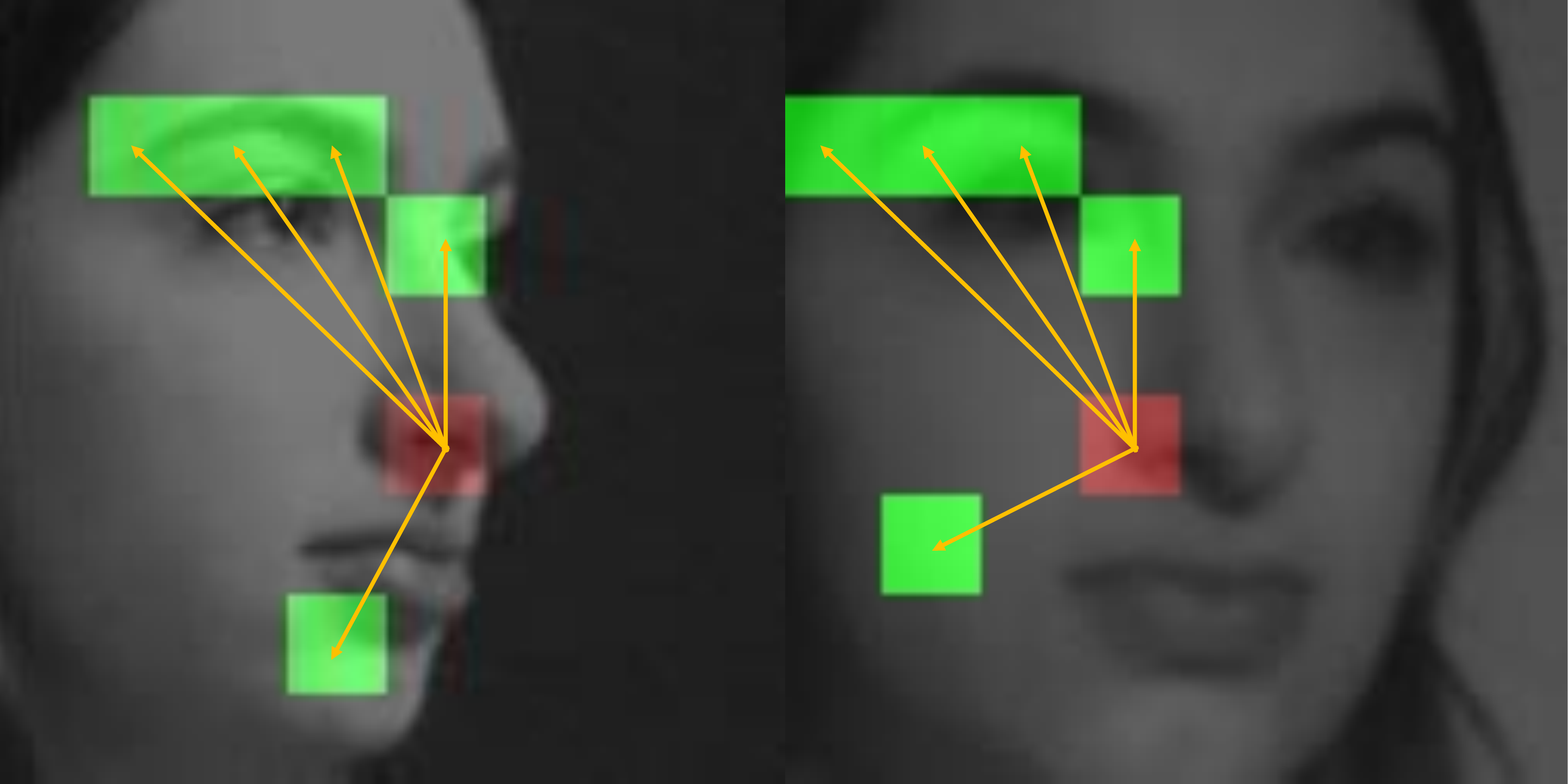}%
		\label{f8_re_c}}
	\hfil
	\subfloat[]{\includegraphics[width=0.22\textwidth]{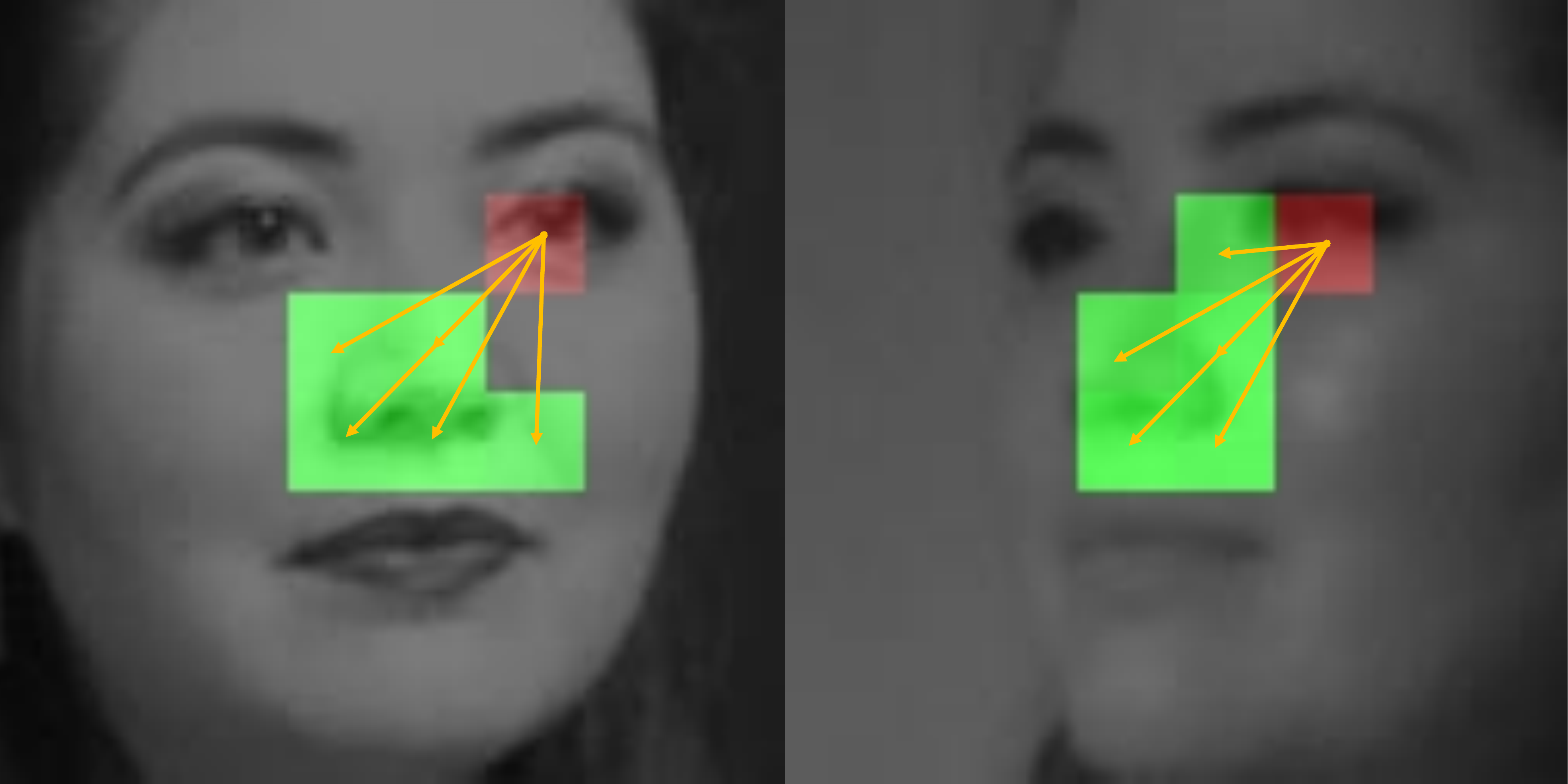}%
		\label{f8_re_d}}
	\hfil
	\\
	\vspace{-0.7\baselineskip}
	\subfloat[]{\includegraphics[width=0.22\textwidth]{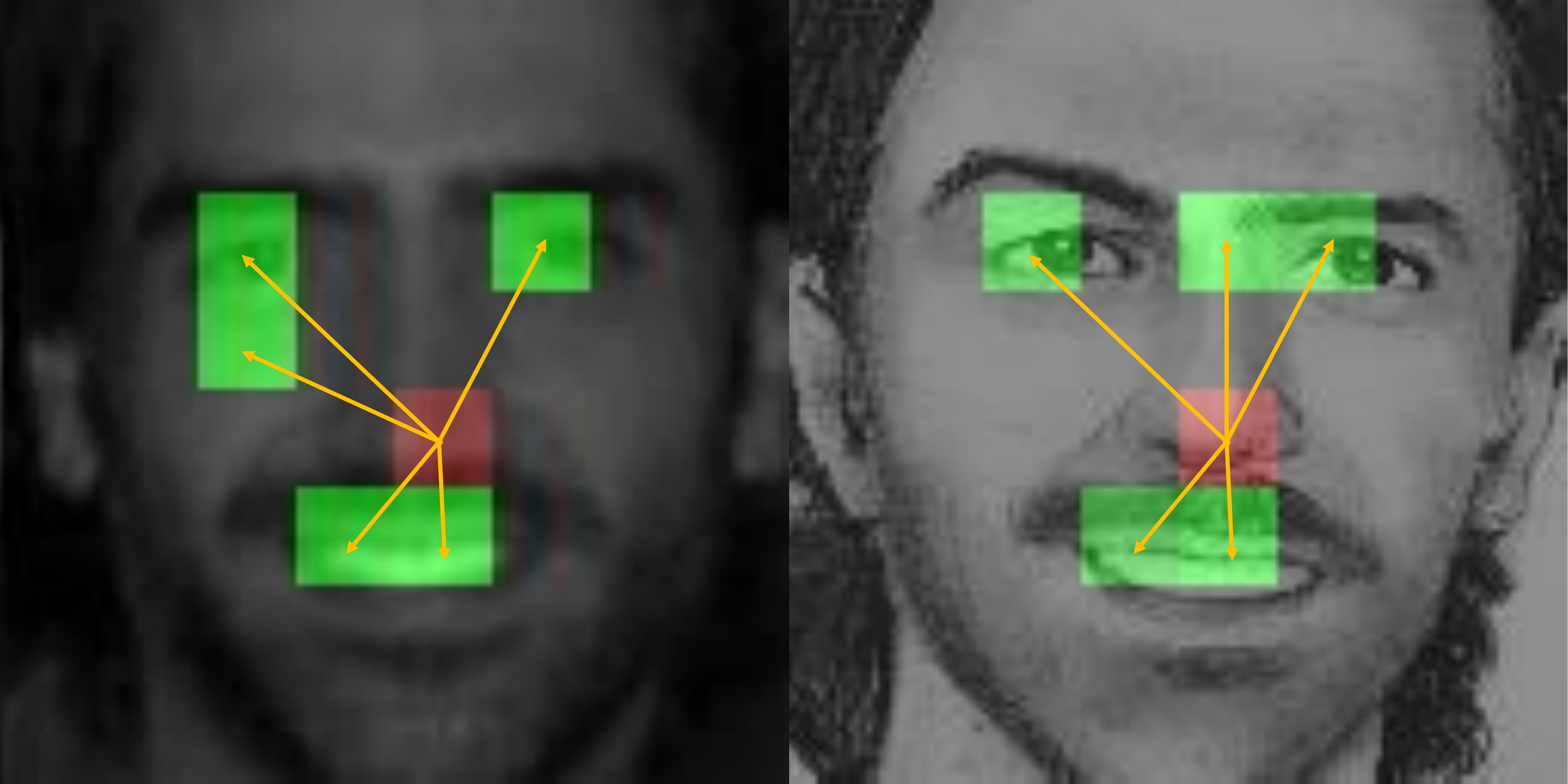}%
		\label{f8_re_e}}
	\hfil
	\subfloat[]{\includegraphics[width=0.22\textwidth]{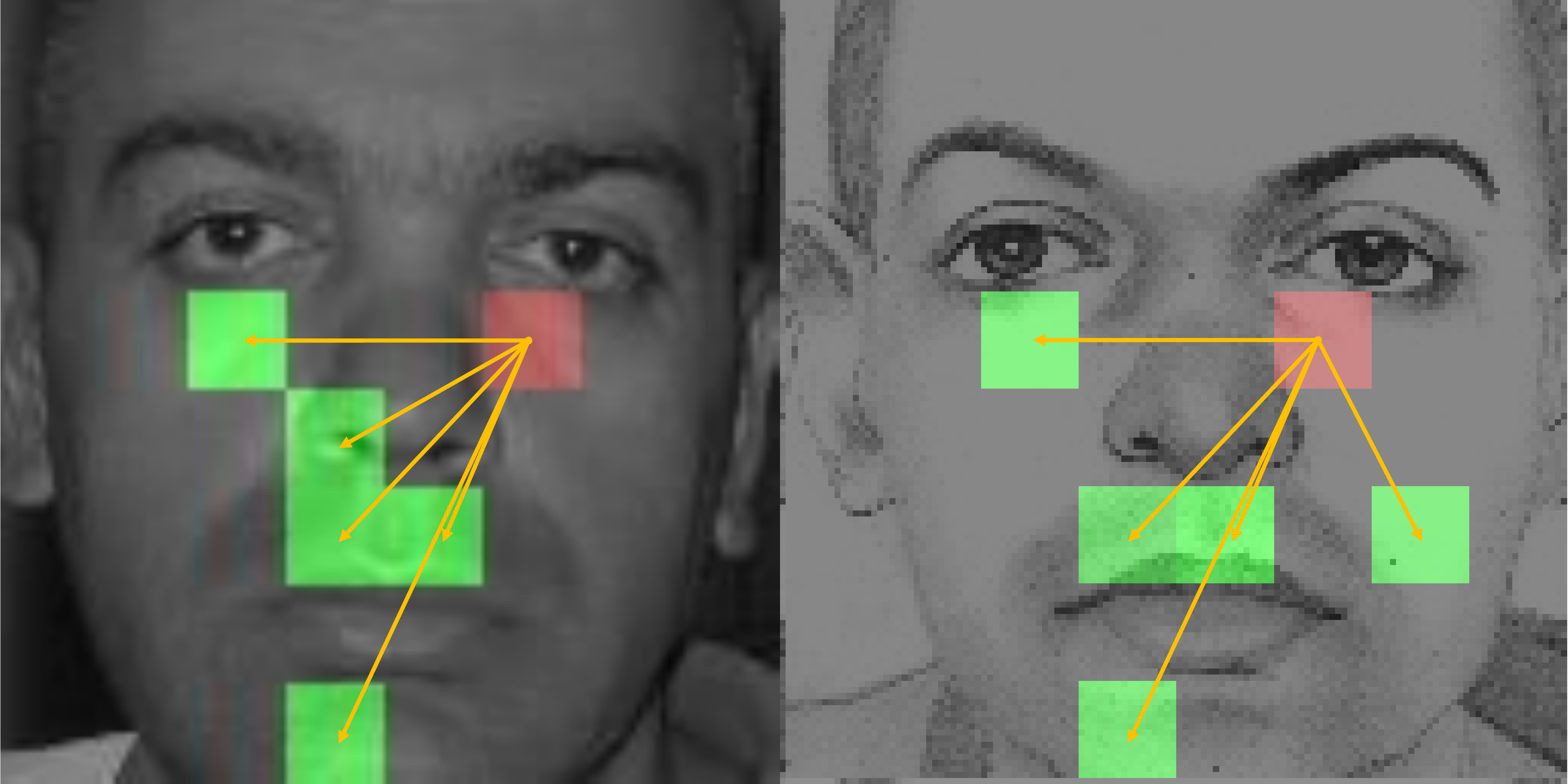}%
		\label{f8_re_f}}
	\hfil
	\subfloat[]{\includegraphics[width=0.22\textwidth]{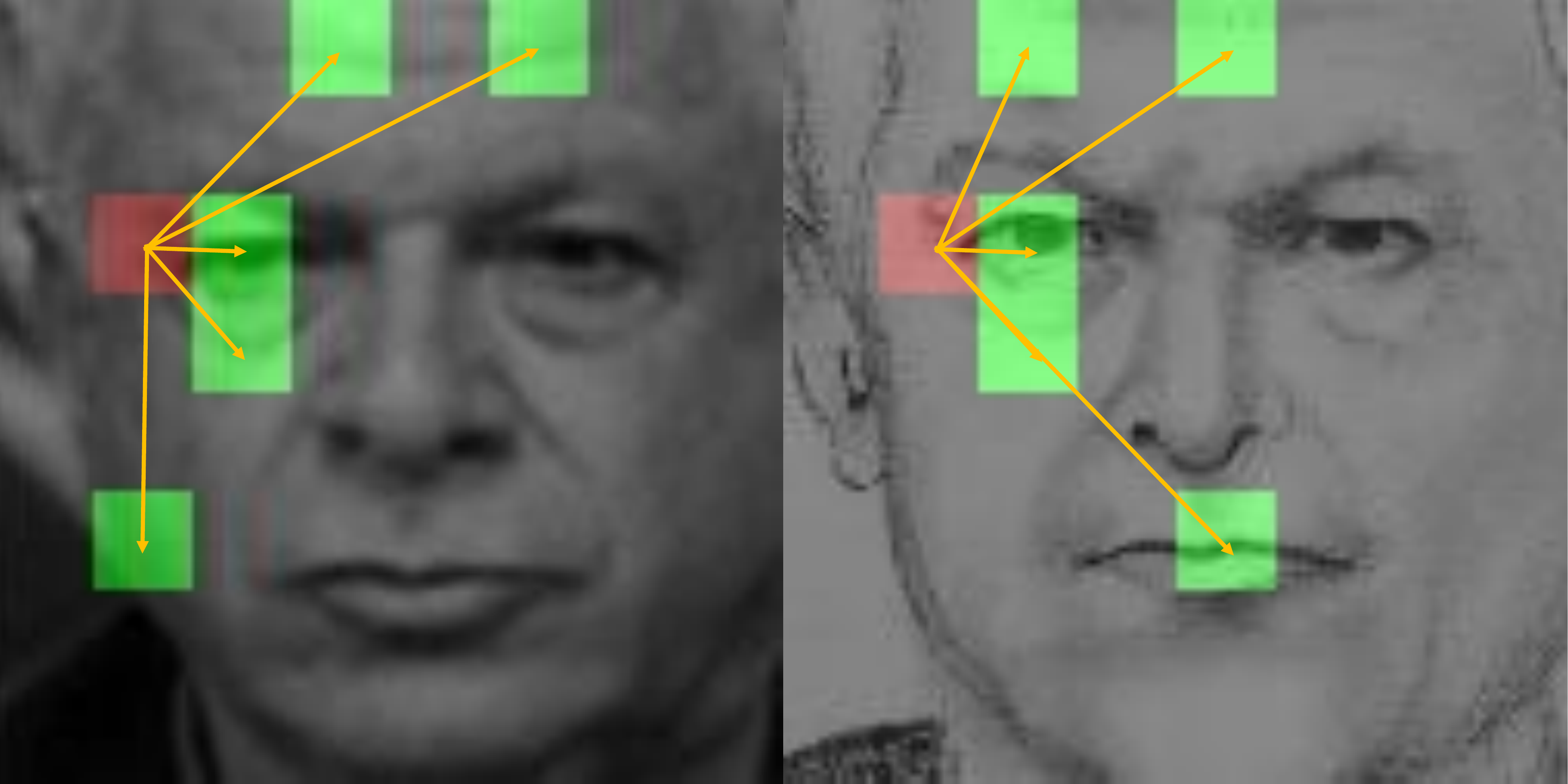}%
		\label{f8_re_g}}
	\hfil
	\subfloat[]{\includegraphics[width=0.22\textwidth]{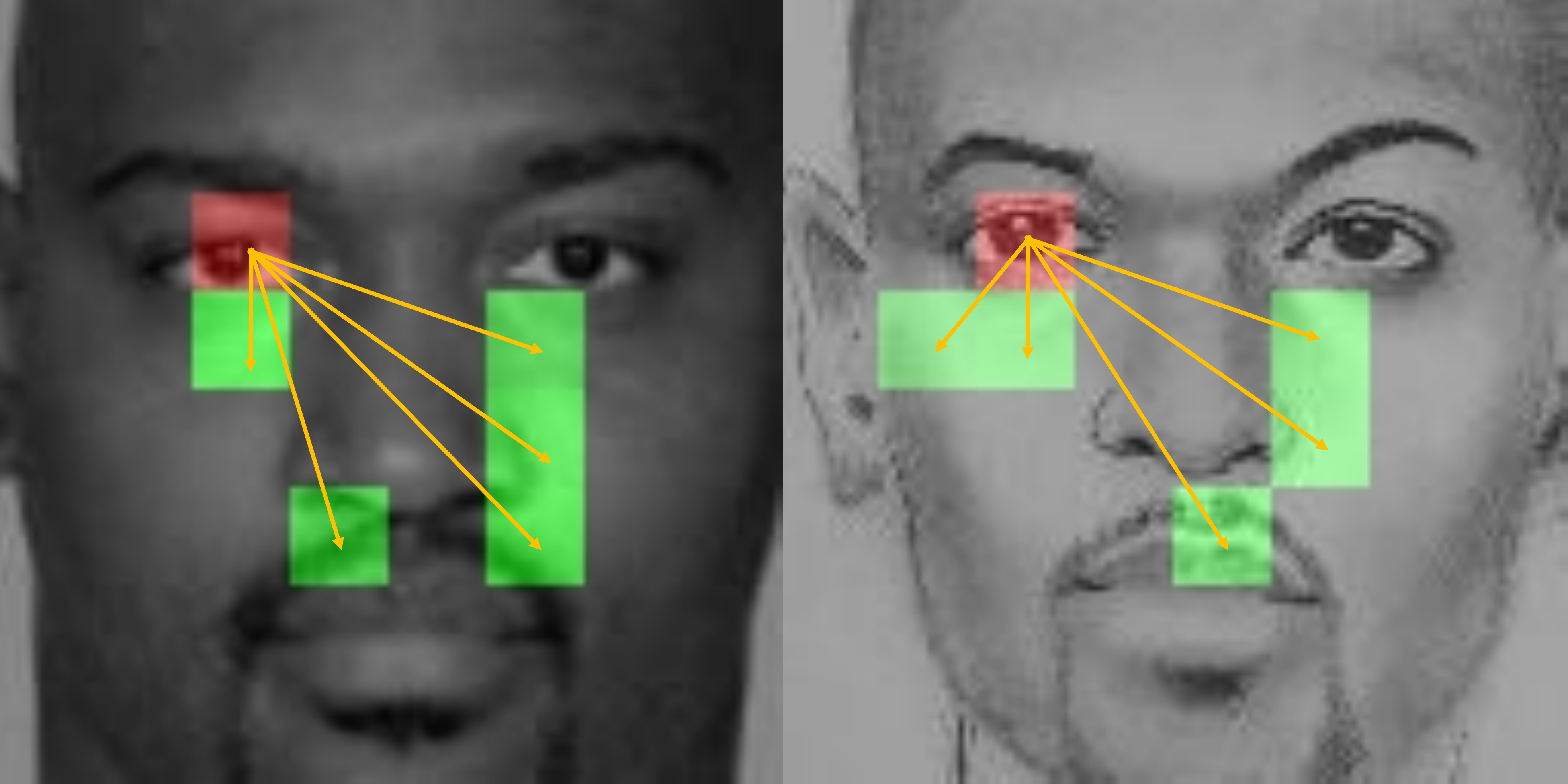}%
		\label{f8_re_h}}
	\caption{Visualization of relational information from RGM edges: (top) a test set from CASIA NIR-VIS 2.0, TUFTS and (bottom) a test set from the IIIT-D Sketch database. In each pair, the right and left sides are the gallery and probe images, respectively. In each image, we select one reference node vector and visualize strong relational node vectors of it. The red region indicates spatial location of $\boldsymbol{n}_{i}$ while green regions indicate spatial location of $\boldsymbol{n}_{j}$ correspond to the top-5 relational edge values $\boldsymbol{A}_{i,j}$ (Equation~\ref{e2}).} 
	\label{f8}
\end{figure*}

\begin{table}[t]
	\begin{center}
	\caption{Comparison with other methods on the BUAA Vis-Nir database}
		\renewcommand{\arraystretch}{1}
		\begin{tabular}{ccc}
			\hline
			Model & \multicolumn{1}{c}{Rank-1 \newline{}Acc(\%)} & \multicolumn{1}{c}{VR@FAR\newline{}=1\%(\%)} \\
			\hline
			\hline
			H2 (LBP3)\cite{shao2016cross} & 88.8  & 88.8 \\
			TRIVLET\cite{liu2016transferring} & 93.9  & 93 \\
			ADFL\cite{song2018adversarial} & 95.2  & 95.3 \\
			CDL\cite{wu2018coupled} & 96.9  & 95.9 \\
			WCNN\cite{he2018wasserstein} & 97.4  & 96 \\
			\textbf{Ours}  & \textbf{99.67} & \textbf{99.22} \\
			\hline
	\end{tabular}
	\end{center}
	\label{t7}%
\vspace{-4mm}
\end{table}%

\subsection{BUAA-VisNir}
\subsubsection{Database}
The BUAA-VisNir database is composed of NIR and VIS face images of 150 subjects. Each subject has nine NIR and VIS images including one frontal view, four different other views, and four different expressions (happiness, anger, disgust and amazement). 
These NIR and VIS images are paired and captured simultaneously. The training set comprises 50 subjects with 900 images. For testing, 100 subjects with one VIS image each make up the gallery set, with 900 NIR images in the probe set. 

\subsubsection{Ablation Studies}
In Table \ref{t5}, the performance with LightCNN-9 and 29 incrementally improves over baseline when the RGM module, NAU, and conditional margin loss $C$-softmax are added. 
With $C$-softmax loss, the performance improves 2.45\% and 0.11\% respectively. On the other hand, when NAU is added to the ResNet18 baseline, the performance decreases because it becomes more difficult to learn the global node correlation with fewer training set subjects. Our approach brings performance improvements 2.78\%, 1.55\%, and 2.23\% over fine-tune in the three baselines.

\subsubsection{Comparison with Other Methods}
Table~\ref{t7} compares our method with three other types of method (projection based, synthesis based and domain-invariant base method) on H2 (LBP3)\cite{shao2016cross}, TRIVLET\cite{liu2016transferring}, ADFL\cite{song2018adversarial}, CDL\cite{wu2018coupled} and WCNN\cite{he2018wasserstein}. Our method shows better performance than other domain-invariant feature methods such as WCNN, TRIVLET that focus on features themselves rather than relationships.

\subsection{Small-scale HFR database}
\subsubsection{Database}
Oulu-CASIA NIR-VIS facial expression database\cite{zhao2011facial} consists of 80 subjects with six different expressions 
and three different illumination conditions. 
Following the protocols in \cite{he2018wasserstein}, we randomly selected 40 subjects and eight images from each NIR and VIS domain for six expressions. The train set and test set are 20 subjects each. For test set, all 960 number of VIS and NIR images of the 20 subjects are used as gallery and probe set. 

TUFTS NIR and VIS database \cite{panetta2018comprehensive} consists of 100 subjects with large pose variations. The number of images per subject is less than 36, and each subject is photographed at nine equidistant positions around a semicircle with a fixed viewpoint. Since there are no protocols or comparison papers for training and test settings, we cropped all faces to 144x144 and use identities 1–25 as the test set and identities 26–100 as the training set. 
\begin{table}[t]
	\centering
	\caption{Comparison with other angular margin losses}
	\resizebox{\linewidth}{!}{
		\renewcommand{\arraystretch}{1.2}
		\begin{tabular}{ccccc}
			\hline
			\multicolumn{1}{c}{\multirow{2}[1]{*}{Loss}} & \multicolumn{2}{c}{CASIA NIR-VIS 2.0} & \multicolumn{1}{c}{CASIA WebFace} & \multicolumn{1}{c}{Small-CASIA WebFace} \\
			& \multicolumn{1}{c}{Rank-1 (\%)} & \multicolumn{1}{c}{VR@FAR(1\%)} & \multicolumn{1}{c}{LFW Ver(\%)} & \multicolumn{1}{c}{LFW Ver(\%)}\\
			\hline
			\hline 
			Normalized-softmax    & 97.2  & 98.76 & 99.13 (99.1) & 98.55 \\
			A-softmax\cite{liu2017sphereface} & 87.06 & 94.78 &  99.18 (99.11) & 98.32 \\
			CosFace\cite{wang2018cosface} & 97.17 & 99.05 & 99.52 (99.51) & 99.02 \\
			ArcFace\cite{deng2019arcface} & 97.95 & \textbf{99.29} & 99.43 (99.53) & 99.03 \\
			\textbf{C-softmax(Ours)} & \textbf{98.03} & 99.15 & \textbf{99.58} & \textbf{99.27} \\
			\hline
	\end{tabular}}
	\label{t9}%
\vspace{-2mm}
\end{table}%
\subsubsection{Results}
The Figure~\ref{f8}(c) and \ref{f8}(d) show visualization of relationship of face components in different poses in TUFTS database. In the figure, regardless of domain and pose, the strong relational components are similar within subject. The detailed explanation of the visualization will be provided in Section~\ref{v}.

Since Oulu-CASIA and TUFTS databases have too small identities (number of 20 and 25) which performances are 100\% and 99.66\% rank-1 accuracy respectively, we only conduct experiments with ablation studies or visualization. The performance on Oulu-CASIA database increases as our method is added, but saturated (Table \ref{t5}). This is because it has fewer subjects, and it has multiple gallery images per subject, unlike other HFR databases which have only one gallery image per subject.
\begin{figure*}[!t]
	\centering
	\begin{tabular}{ccc}
		\subfloat[ID 1]{\includegraphics[width=0.455\textwidth]{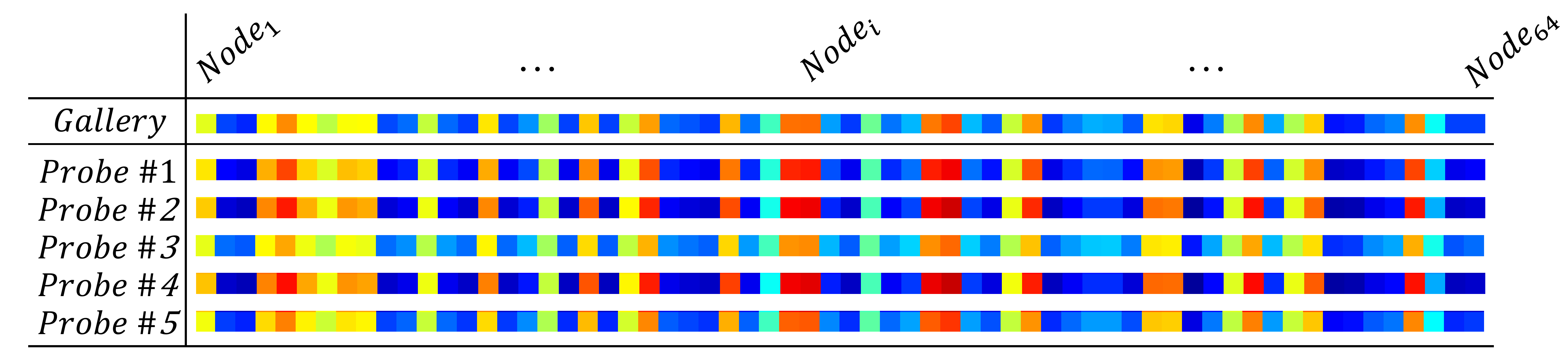}} & 
		\subfloat[ID 2]{\includegraphics[width=0.455\textwidth]{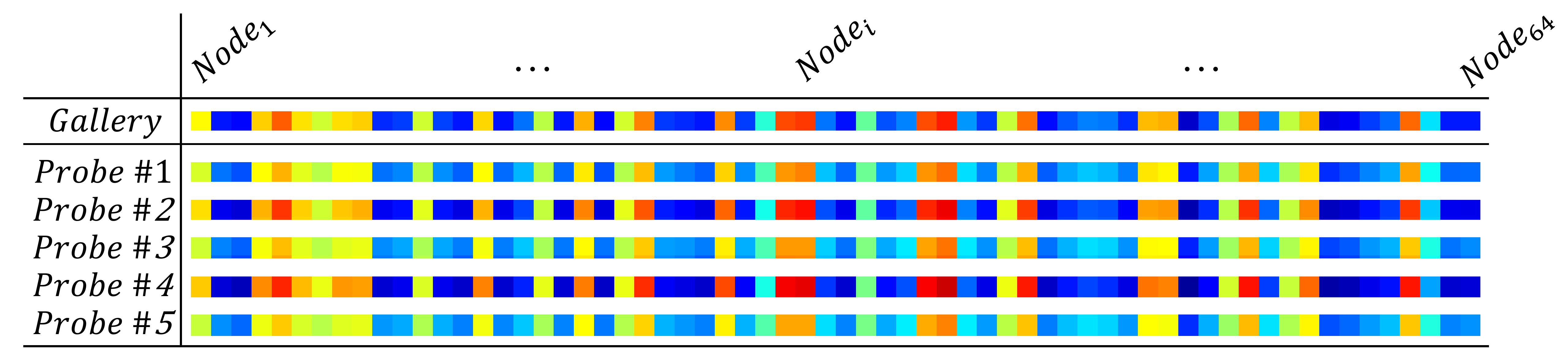}} & 
		\\
		\subfloat[ID 3]{\includegraphics[width=0.455\textwidth]{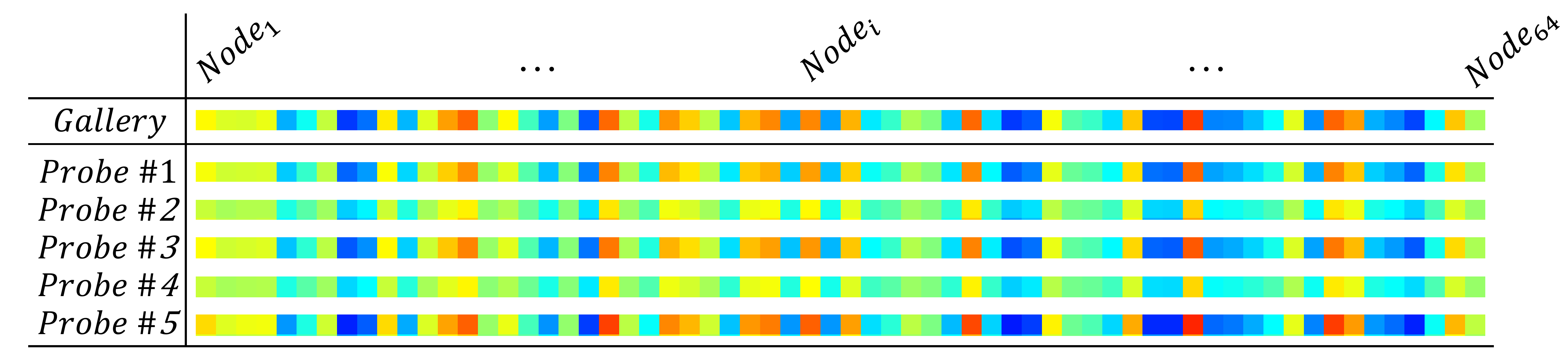}} & 
		\subfloat[ID 4]{\includegraphics[width=0.455\textwidth]{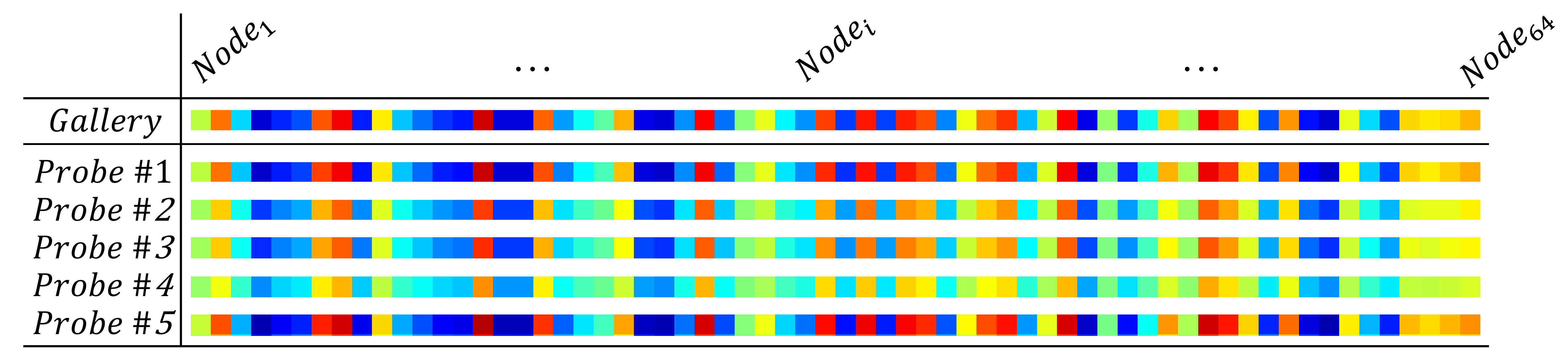}}&
		\multirow{-10.5}[2]{*}{\subfloat{\includegraphics[width=0.045\textwidth]{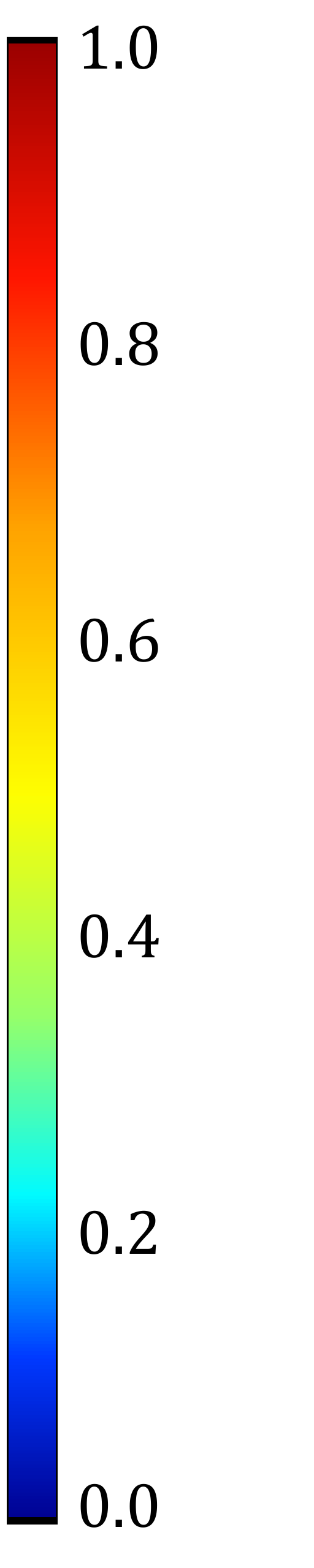}}} \\
	\end{tabular}
	\caption{Visualization of NAU. Each subfigure, row represents randomly selected samples (1 gallery and 5 probes within a same subject) from the CASIA NIR-VIS 2.0 database and column represents node. We extract each node's scalar weighting values $s_{i}$ (see Equation~\ref{e4}) from the NAU after RGM propagation. Regardless of domain, $s_{i}$ is similar in each subject which indicates that patterns of the node importance within a same subject are similar.}
	\label{f9}
\vspace{-2mm}
\end{figure*}
\subsection{Conditional Margin Loss: $C$-softmax} \label{exp loss}
\subsubsection{CASIA NIR-VIS 2.0}
We compare our conditional margin loss ($C$-softmax) to other angular margin losses such as Normalized-softmax\cite{ranjan2017l2, wang2017normface}, A-softmax\cite{liu2017sphereface}, CosFace\cite{wang2018cosface} and ArcFace\cite{deng2019arcface} using LightCNN-9 as a baseline and training under the same conditions (the embeddings are normalized and scaled $s$ = 24); the margin for each loss follows each study. In Table~\ref{t9}, on the CASIA NIR-VIS 2.0 database, performance of ArcFace and our C-softmax show better performance than the other losses at 97.97\% and 98.03\% because of the different margins for the class cosine similarity, as shown in Figure~\ref{f6}. ArcFace reduces the margin when the class cosine similarity is large or small and increases it near the midpoint (Figure~\ref{f6}(b)), while $C$-softmax increases the margin at larger class similarity values (Figure~\ref{f6}(c)). This helps to control classes with domain discrepancy because it effectively adjusts the margin between inter-classes.

\subsubsection{LFW}
In Table~\ref{t9}, we also conduct experiments with LFW \cite{bhatt2012memetic}, a large-scale visual face database. For this purpose, we use CASIA WebFace \cite{yi2014learning} consisting of 10,575 subjects for the training dataset and perform evaluation in LFW. All the implementation details are same as ArcFace and the margin and scale factors of each loss function are set to the optimal values presented in the original paper. The performance of each loss function is reproduced and the value in parentheses is the performance written on each paper. The results show that $C$-softmax achieves best performance with 99.58\%, followed by the CosFace and ArcFace. In addition, we conduct experiments on a small-CASIA WebFace database with 5,287 subjects, half the size of the CASIA WebFace. The performance gap between $C$-softmax and other losses is larger in small-CASIA WebFace than in CASIA WebFace. This result shows that $C$-softmax improves training effectiveness when we train on the dataset with a small number of classes.
\begin{table}[t]
	\centering
	\caption{Comparison of softmax and Sigmoid activation in RGM}
	\resizebox{\linewidth}{!}{
		\renewcommand{\arraystretch}{1.1}
		\begin{tabular}{cccccc}
			\hline
			\multicolumn{1}{c}{\multirow{2}[1]{*}{Database}}&\multicolumn{1}{p{4.05em}}{Activation \newline{}function}& \multicolumn{1}{p{4.05em}}{Rank-1 \newline{}Acc(\%)} & \multicolumn{1}{p{4.05em}}{VR@FAR\newline{}=1\%(\%)} & \multicolumn{1}{p{4.05em}}{VR@FAR\newline{}=0.1\%(\%)} & \multicolumn{1}{p{4.05em}}{VR@FAR\newline{}=0.01\%(\%)} \\
			\hline
			\hline
			\multicolumn{1}{c}{\multirow{2}[2]{*}{CASIA NIR}} & softmax & 97.95 & \textbf{99.16} & 97.15 & \textbf{96} \\
			& \textbf{sigmoid} & \textbf{98.03} & 99.15 & \textbf{96.76} & 95.23 \\
			\hline
			\multicolumn{1}{c}{\multirow{2}[2]{*}{IIITD \newline Sketch}} & softmax & 87.66 & 95.32 & 90.21 & - \\
			& \textbf{sigmoid} & \textbf{88.51} & \textbf{96.17} & \textbf{94.47} & - \\
			\hline
	\end{tabular}}
	\label{t10}%
\vspace{-3mm}
\end{table}%
\subsection{Discussion}
\subsubsection{RGM with Sigmoid Activation Function} \label{sigmoid}
As mentioned in Section~\ref{RGM}, when obtaining a directed relation between nodes, all edge values are passed through an activation function. In this case, we use a sigmoid activation function instead of softmax because relation information for each node is independent of and should not be influenced by other nodes' relations. Unlike sigmoids where $f_{sigmoid}(\boldsymbol{x}_{i}) = {1}/({1+e^{-\boldsymbol{x}_{i}}})$, softmax $f_{softmax}(\boldsymbol{x}_{i}) = {\boldsymbol{x}_{i}}/{\sum e^{-\boldsymbol{x}_{j}}}$ looks at the interrelation of all values. Table~\ref {t10} shows the results of experiments in which the activation function of RGM is varied. We use LightCNN-9 as a baseline with the CASIA NIR-VIS and IIIT-D Sketch databases. For these databases, the rank-1 accuracy is increased by 0.08\% and 0.85\% compared to softmax.

\subsubsection{Visualization of Relations} \label{v}
We visualize node relation $\boldsymbol{A}_{i,j}$ in Equation \ref{e2} which is the directed relations of nodes in the RGM. In Figure~\ref{f8}, the first row shows VIS and NIR pair, while the second row shows VIS and Sketch pairs. In each image, we select one reference node vector (red color) and visualize five strong relational node vectors (green color) of it. Regardless of domain, within a subject, the strong relational components in the gallery and probe images are similar. For example, the subject in Figure \ref{f8}(a), the left eye has a strong relationship with right eye, nose and wrinkles; the subject in Figure \ref{f8}(b), the left eye has a strong relationship with wrinkles, nose, and mouth; In the case of different subjects, when comparing (a) and (b), the strong relationship components with the left eye are different. Also, in case of pose variation, both gallery and probe in Figure \ref{f8}(c), nostrils have a strong relationship with left eyebrow and eye, whereas right eye has a strong relationship with nose in Figure \ref{f8}(d). These relationships are obtained by passing the gallery VIS image and the probe NIR or Sketch image separately to the RGM, revealing each identity has similar relationships in faces regardless of domain. Additional visualization results are presented in the Supplementary Material.

\subsubsection{Visualization of Node Attention Unit}
Nodes whose relational information is propagated through the RGM are node-wisely recalibrated through the NAU. Figure~\ref{f9} shows the scale value computed for node-wise recalibration from the NAU ($s$ in Equation~\ref{e4}). Each row corresponds to samples of four subjects from test set; in each subject, the first row is a gallery and other rows are probe sets. Looking at the gallery and probe set, we can observe that the nodes are similarly focused for each subject. In other words, regardless of domain, the importance pattern of relation propagated nodes (output of RGM) are different across subjects and similar within each subject. 


\section{Conclusion} \label{conclusion}
The Relational Graph Module (RGM) extracts representative relational information of each identity by embedding each face component into a node vector and modeling the relationships among these. This graph-structured module solved the discrepancy problem between HFR domains using a structured approach based on extracting relations. Moreover, the RGM overcame the problem of lack of adequate HFR database by plugging into a pre-trained face extractor and fine-tuning it. In addition, through the Node Attention Unit (NAU), node-wise recalibration was performed to focus on global informative nodes among propagated node vectors. Furthermore, our novel $C$-softmax loss helped to learn common projection space adaptively by applying a higher margin as the class similarity increases. 

We applied the RGM module to several pre-trained backbones and explored performance improvements on NIR-to-VIS and Sketch-to-VIS tasks. In addition, in ablation studies, each proposed method showed the impact of its role through boosted performance. Furthermore, the visualization of relational information in VIS, NIR, and sketch images showed that relationships within the face are similar in each subject, revealing representative domain-invariant features. Finally, our proposed approach showed best performance comparing with state-of-the-art methods on the CASIA NIR-VIS 2.0, IIIT-D Sketch, BUAA-VisNir, Oulu-CASIA NIR-VIS and TUFTS.

{
 \bibliographystyle{IEEEtran}
 \bibliography{egbib}

\begin{thebibliography}{10}
\providecommand{\url}[1]{#1}
\csname url@samestyle\endcsname
\providecommand{\newblock}{\relax}
\providecommand{\bibinfo}[2]{#2}
\providecommand{\BIBentrySTDinterwordspacing}{\spaceskip=0pt\relax}
\providecommand{\BIBentryALTinterwordstretchfactor}{4}
\providecommand{\BIBentryALTinterwordspacing}{\spaceskip=\fontdimen2\font plus
\BIBentryALTinterwordstretchfactor\fontdimen3\font minus
  \fontdimen4\font\relax}
\providecommand{\BIBforeignlanguage}[2]{{%
\expandafter\ifx\csname l@#1\endcsname\relax
\typeout{** WARNING: IEEEtran.bst: No hyphenation pattern has been}%
\typeout{** loaded for the language `#1'. Using the pattern for}%
\typeout{** the default language instead.}%
\else
\language=\csname l@#1\endcsname
\fi
#2}}
\providecommand{\BIBdecl}{\relax}
\BIBdecl

\bibitem{wang2018deep}
M.~Wang and W.~Deng, ``Deep face recognition: A survey,'' \emph{arXiv preprint
  arXiv:1804.06655}, 2018.

\bibitem{sengupta2016frontal}
S.~Sengupta, J.-C. Chen, C.~Castillo, V.~M. Patel, R.~Chellappa, and D.~W.
  Jacobs, ``Frontal to profile face verification in the wild,'' in \emph{2016
  IEEE Winter Conference on Applications of Computer Vision (WACV)}.\hskip 1em
  plus 0.5em minus 0.4em\relax IEEE, 2016, pp. 1--9.

\bibitem{wang2018face}
Z.~Wang, X.~Tang, W.~Luo, and S.~Gao, ``Face aging with identity-preserved
  conditional generative adversarial networks,'' in \emph{Proceedings of the
  IEEE Conference on Computer Vision and Pattern Recognition}, 2018, pp.
  7939--7947.

\bibitem{zangeneh2020low}
E.~Zangeneh, M.~Rahmati, and Y.~Mohsenzadeh, ``Low resolution face recognition
  using a two-branch deep convolutional neural network architecture,''
  \emph{Expert Systems with Applications}, vol. 139, p. 112854, 2020.

\bibitem{he2017learning}
R.~He, X.~Wu, Z.~Sun, and T.~Tan, ``Learning invariant deep representation for
  nir-vis face recognition,'' in \emph{Thirty-First AAAI Conference on
  Artificial Intelligence}, 2017.

\bibitem{bi2019multi}
H.~Bi, N.~Li, H.~Guan, D.~Lu, and L.~Yang, ``A multi-scale conditional
  generative adversarial network for face sketch synthesis,'' in \emph{2019
  IEEE International Conference on Image Processing (ICIP)}.\hskip 1em plus
  0.5em minus 0.4em\relax IEEE, 2019, pp. 3876--3880.

\bibitem{ouyang2016survey}
S.~Ouyang, T.~Hospedales, Y.-Z. Song, X.~Li, C.~C. Loy, and X.~Wang, ``A survey
  on heterogeneous face recognition: Sketch, infra-red, 3d and
  low-resolution,'' \emph{Image and Vision Computing}, vol.~56, pp. 28--48,
  2016.

\bibitem{li2013casia}
S.~Li, D.~Yi, Z.~Lei, and S.~Liao, ``The casia nir-vis 2.0 face database,'' in
  \emph{Proceedings of the IEEE conference on computer vision and pattern
  recognition workshops}, 2013, pp. 348--353.

\bibitem{bhatt2012memetic}
H.~S. Bhatt, S.~Bharadwaj, R.~Singh, and M.~Vatsa, ``Memetic approach for
  matching sketches with digital face images,'' Tech. Rep., 2012.

\bibitem{huang2012buaa}
D.~Huang, J.~Sun, and Y.~Wang, ``The buaa-visnir face database instructions,''
  \emph{School Comput. Sci. Eng., Beihang Univ., Beijing, China, Tech. Rep.
  IRIP-TR-12-FR-001}, 2012.

\bibitem{guo2016ms}
Y.~Guo, L.~Zhang, Y.~Hu, X.~He, and J.~Gao, ``Ms-celeb-1m: A dataset and
  benchmark for large-scale face recognition,'' in \emph{European Conference on
  Computer Vision}.\hskip 1em plus 0.5em minus 0.4em\relax Springer, 2016, pp.
  87--102.

\bibitem{kemelmacher2016megaface}
I.~Kemelmacher-Shlizerman, S.~M. Seitz, D.~Miller, and E.~Brossard, ``The
  megaface benchmark: 1 million faces for recognition at scale,'' in
  \emph{Proceedings of the IEEE Conference on Computer Vision and Pattern
  Recognition}, 2016, pp. 4873--4882.

\bibitem{song2018adversarial}
L.~Song, M.~Zhang, X.~Wu, and R.~He, ``Adversarial discriminative heterogeneous
  face recognition,'' in \emph{Thirty-Second AAAI Conference on Artificial
  Intelligence}, 2018.

\bibitem{zhang2018tv}
T.~Zhang, A.~Wiliem, S.~Yang, and B.~Lovell, ``Tv-gan: Generative adversarial
  network based thermal to visible face recognition,'' in \emph{2018
  international conference on biometrics (ICB)}.\hskip 1em plus 0.5em minus
  0.4em\relax IEEE, 2018, pp. 174--181.

\bibitem{he2018wasserstein}
R.~He, X.~Wu, Z.~Sun, and T.~Tan, ``Wasserstein cnn: Learning invariant
  features for nir-vis face recognition,'' \emph{IEEE transactions on pattern
  analysis and machine intelligence}, vol.~41, no.~7, pp. 1761--1773, 2018.

\bibitem{wu2018coupled}
X.~Wu, L.~Song, R.~He, and T.~Tan, ``Coupled deep learning for heterogeneous
  face recognition,'' in \emph{Thirty-Second AAAI Conference on Artificial
  Intelligence}, 2018.

\bibitem{liu2016transferring}
X.~Liu, L.~Song, X.~Wu, and T.~Tan, ``Transferring deep representation for
  nir-vis heterogeneous face recognition,'' in \emph{2016 International
  Conference on Biometrics (ICB)}.\hskip 1em plus 0.5em minus 0.4em\relax IEEE,
  2016, pp. 1--8.

\bibitem{deng2019residual}
Z.~Deng, X.~Peng, and Y.~Qiao, ``Residual compensation networks for
  heterogeneous face recognition,'' in \emph{Proceedings of the AAAI Conference
  on Artificial Intelligence}, vol.~33, 2019, pp. 8239--8246.

\bibitem{myeong2019rm}
M.Cho, T.~Chung, T.~Kim, and S.~Lee, ``Nir-to-vis face recognition via
  embedding relations and coordinates of the pairwise features,'' in \emph{2019
  international conference on biometrics (ICB)}.\hskip 1em plus 0.5em minus
  0.4em\relax IEEE, 2019.

\bibitem{chowdhury2016one}
A.~R. Chowdhury, T.-Y. Lin, S.~Maji, and E.~Learned-Miller, ``One-to-many face
  recognition with bilinear cnns,'' in \emph{2016 IEEE Winter Conference on
  Applications of Computer Vision (WACV)}.\hskip 1em plus 0.5em minus
  0.4em\relax IEEE, 2016, pp. 1--9.

\bibitem{chen20182}
Y.~Chen, Y.~Kalantidis, J.~Li, S.~Yan, and J.~Feng, ``A\^{} 2-nets: Double
  attention networks,'' in \emph{Advances in Neural Information Processing
  Systems}, 2018, pp. 352--361.

\bibitem{wang2018non}
X.~Wang, R.~Girshick, A.~Gupta, and K.~He, ``Non-local neural networks,'' in
  \emph{Proceedings of the IEEE Conference on Computer Vision and Pattern
  Recognition}, 2018, pp. 7794--7803.

\bibitem{santoro2017simple}
A.~Santoro, D.~Raposo, D.~G. Barrett, M.~Malinowski, R.~Pascanu, P.~Battaglia,
  and T.~Lillicrap, ``A simple neural network module for relational
  reasoning,'' in \emph{Advances in neural information processing systems},
  2017, pp. 4967--4976.

\bibitem{lin2006inter}
D.~Lin and X.~Tang, ``Inter-modality face recognition,'' in \emph{European
  conference on computer vision}.\hskip 1em plus 0.5em minus 0.4em\relax
  Springer, 2006, pp. 13--26.

\bibitem{yi2009partial}
D.~Yi, S.~Liao, Z.~Lei, J.~Sang, and S.~Z. Li, ``Partial face matching between
  near infrared and visual images in mbgc portal challenge,'' in
  \emph{International Conference on Biometrics}.\hskip 1em plus 0.5em minus
  0.4em\relax Springer, 2009, pp. 733--742.

\bibitem{lei2009coupled}
Z.~Lei and S.~Z. Li, ``Coupled spectral regression for matching heterogeneous
  faces,'' in \emph{2009 IEEE Conference on Computer Vision and Pattern
  Recognition}.\hskip 1em plus 0.5em minus 0.4em\relax IEEE, 2009, pp.
  1123--1128.

\bibitem{lei2012coupled}
Z.~Lei, S.~Liao, A.~K. Jain, and S.~Z. Li, ``Coupled discriminant analysis for
  heterogeneous face recognition,'' \emph{IEEE Transactions on Information
  Forensics and Security}, vol.~7, no.~6, pp. 1707--1716, 2012.

\bibitem{shao2014generalized}
M.~Shao, D.~Kit, and Y.~Fu, ``Generalized transfer subspace learning through
  low-rank constraint,'' \emph{International Journal of Computer Vision}, vol.
  109, no. 1-2, pp. 74--93, 2014.

\bibitem{sarfraz2015deep}
M.~S. Sarfraz and R.~Stiefelhagen, ``Deep perceptual mapping for thermal to
  visible face recognition,'' \emph{arXiv preprint arXiv:1507.02879}, 2015.

\bibitem{reale2016seeing}
C.~Reale, N.~M. Nasrabadi, H.~Kwon, and R.~Chellappa, ``Seeing the forest from
  the trees: A holistic approach to near-infrared heterogeneous face
  recognition,'' in \emph{Proceedings of the IEEE Conference on Computer Vision
  and Pattern Recognition Workshops}, 2016, pp. 54--62.

\bibitem{liu2005nonlinear}
Q.~Liu, X.~Tang, H.~Jin, H.~Lu, and S.~Ma, ``A nonlinear approach for face
  sketch synthesis and recognition,'' in \emph{2005 IEEE Computer Society
  Conference on Computer Vision and Pattern Recognition (CVPR'05)},
  vol.~1.\hskip 1em plus 0.5em minus 0.4em\relax IEEE, 2005, pp. 1005--1010.

\bibitem{wang2008face}
X.~Wang and X.~Tang, ``Face photo-sketch synthesis and recognition,''
  \emph{IEEE Transactions on Pattern Analysis and Machine Intelligence},
  vol.~31, no.~11, pp. 1955--1967, 2008.

\bibitem{goodfellow2014generative}
I.~Goodfellow, J.~Pouget-Abadie, M.~Mirza, B.~Xu, D.~Warde-Farley, S.~Ozair,
  A.~Courville, and Y.~Bengio, ``Generative adversarial nets,'' in
  \emph{Advances in neural information processing systems}, 2014, pp.
  2672--2680.

\bibitem{zhu2017unpaired}
J.-Y. Zhu, T.~Park, P.~Isola, and A.~A. Efros, ``Unpaired image-to-image
  translation using cycle-consistent adversarial networks,'' in
  \emph{Proceedings of the IEEE international conference on computer vision},
  2017, pp. 2223--2232.

\bibitem{liu2012heterogeneous}
S.~Liu, D.~Yi, Z.~Lei, and S.~Z. Li, ``Heterogeneous face image matching using
  multi-scale features,'' in \emph{2012 5th IAPR International Conference on
  Biometrics (ICB)}.\hskip 1em plus 0.5em minus 0.4em\relax IEEE, 2012, pp.
  79--84.

\bibitem{schroff2015facenet}
F.~Schroff, D.~Kalenichenko, and J.~Philbin, ``Facenet: A unified embedding for
  face recognition and clustering,'' in \emph{Proceedings of the IEEE
  conference on computer vision and pattern recognition}, 2015, pp. 815--823.

\bibitem{wu2019disentangled}
X.~Wu, H.~Huang, V.~M. Patel, R.~He, and Z.~Sun, ``Disentangled variational
  representation for heterogeneous face recognition,'' in \emph{Proceedings of
  the AAAI Conference on Artificial Intelligence}, vol.~33, 2019, pp.
  9005--9012.

\bibitem{klare2012heterogeneous}
B.~F. Klare and A.~K. Jain, ``Heterogeneous face recognition using kernel
  prototype similarities,'' \emph{IEEE transactions on pattern analysis and
  machine intelligence}, vol.~35, no.~6, pp. 1410--1422, 2012.

\bibitem{peng2016graphical}
C.~Peng, X.~Gao, N.~Wang, and J.~Li, ``Graphical representation for
  heterogeneous face recognition,'' \emph{IEEE transactions on pattern analysis
  and machine intelligence}, vol.~39, no.~2, pp. 301--312, 2016.

\bibitem{peng2019sparse}
------, ``Sparse graphical representation based discriminant analysis for
  heterogeneous face recognition,'' \emph{Signal Processing}, vol. 156, pp.
  46--61, 2019.

\bibitem{lin2015bilinear}
T.-Y. Lin, A.~RoyChowdhury, and S.~Maji, ``Bilinear cnn models for fine-grained
  visual recognition,'' in \emph{Proceedings of the IEEE international
  conference on computer vision}, 2015, pp. 1449--1457.

\bibitem{sun2018actor}
C.~Sun, A.~Shrivastava, C.~Vondrick, K.~Murphy, R.~Sukthankar, and C.~Schmid,
  ``Actor-centric relation network,'' in \emph{Proceedings of the European
  Conference on Computer Vision (ECCV)}, 2018, pp. 318--334.

\bibitem{zhou2018graph}
J.~Zhou, G.~Cui, Z.~Zhang, C.~Yang, Z.~Liu, and M.~Sun, ``Graph neural
  networks: A review of methods and applications,'' \emph{arXiv preprint
  arXiv:1812.08434}, 2018.

\bibitem{kipf2016semi}
T.~N. Kipf and M.~Welling, ``Semi-supervised classification with graph
  convolutional networks,'' \emph{arXiv preprint arXiv:1609.02907}, 2016.

\bibitem{wang2018videos}
X.~Wang and A.~Gupta, ``Videos as space-time region graphs,'' in
  \emph{Proceedings of the European Conference on Computer Vision (ECCV)},
  2018, pp. 399--417.

\bibitem{chen2019graph}
Y.~Chen, M.~Rohrbach, Z.~Yan, Y.~Shuicheng, J.~Feng, and Y.~Kalantidis,
  ``Graph-based global reasoning networks,'' in \emph{Proceedings of the IEEE
  Conference on Computer Vision and Pattern Recognition}, 2019, pp. 433--442.

\bibitem{yi2014learning}
D.~Yi, Z.~Lei, S.~Liao, and S.~Z. Li, ``Learning face representation from
  scratch,'' \emph{arXiv preprint arXiv:1411.7923}, 2014.

\bibitem{velivckovic2017graph}
P.~Veli{\v{c}}kovi{\'c}, G.~Cucurull, A.~Casanova, A.~Romero, P.~Lio, and
  Y.~Bengio, ``Graph attention networks,'' \emph{arXiv preprint
  arXiv:1710.10903}, 2017.

\bibitem{hu2018squeeze}
J.~Hu, L.~Shen, and G.~Sun, ``Squeeze-and-excitation networks,'' in
  \emph{Proceedings of the IEEE conference on computer vision and pattern
  recognition}, 2018, pp. 7132--7141.

\bibitem{woo2018cbam}
S.~Woo, J.~Park, J.-Y. Lee, and I.~So~Kweon, ``Cbam: Convolutional block
  attention module,'' in \emph{Proceedings of the European Conference on
  Computer Vision (ECCV)}, 2018, pp. 3--19.

\bibitem{ranjan2017l2}
R.~Ranjan, C.~D. Castillo, and R.~Chellappa, ``L2-constrained softmax loss for
  discriminative face verification,'' \emph{arXiv preprint arXiv:1703.09507},
  2017.

\bibitem{wang2018cosface}
H.~Wang, Y.~Wang, Z.~Zhou, X.~Ji, D.~Gong, J.~Zhou, Z.~Li, and W.~Liu,
  ``Cosface: Large margin cosine loss for deep face recognition,'' in
  \emph{Proceedings of the IEEE Conference on Computer Vision and Pattern
  Recognition}, 2018, pp. 5265--5274.

\bibitem{deng2019arcface}
J.~Deng, J.~Guo, N.~Xue, and S.~Zafeiriou, ``Arcface: Additive angular margin
  loss for deep face recognition,'' in \emph{Proceedings of the IEEE Conference
  on Computer Vision and Pattern Recognition}, 2019, pp. 4690--4699.

\bibitem{zhao2011facial}
G.~Zhao, X.~Huang, M.~Taini, S.~Z. Li, and M.~Pietik{\"a}Inen, ``Facial
  expression recognition from near-infrared videos,'' \emph{Image and Vision
  Computing}, vol.~29, no.~9, pp. 607--619, 2011.

\bibitem{panetta2018comprehensive}
K.~Panetta, Q.~Wan, S.~Agaian, S.~Rajeev, S.~Kamath, R.~Rajendran, S.~Rao,
  A.~Kaszowska, H.~Taylor, A.~Samani \emph{et~al.}, ``A comprehensive database
  for benchmarking imaging systems,'' \emph{IEEE transactions on pattern
  analysis and machine intelligence}, 2018.

\bibitem{wu2018light}
X.~Wu, R.~He, Z.~Sun, and T.~Tan, ``A light cnn for deep face representation
  with noisy labels,'' \emph{IEEE Transactions on Information Forensics and
  Security}, vol.~13, no.~11, pp. 2884--2896, 2018.

\bibitem{he2016deep}
K.~He, X.~Zhang, S.~Ren, and J.~Sun, ``Deep residual learning for image
  recognition,'' in \emph{Proceedings of the IEEE conference on computer vision
  and pattern recognition}, 2016, pp. 770--778.

\bibitem{srivastava2014dropout}
N.~Srivastava, G.~Hinton, A.~Krizhevsky, I.~Sutskever, and R.~Salakhutdinov,
  ``Dropout: a simple way to prevent neural networks from overfitting,''
  \emph{The journal of machine learning research}, vol.~15, no.~1, pp.
  1929--1958, 2014.

\bibitem{saxena2016heterogeneous}
S.~Saxena and J.~Verbeek, ``Heterogeneous face recognition with cnns,'' in
  \emph{European conference on computer vision}.\hskip 1em plus 0.5em minus
  0.4em\relax Springer, 2016, pp. 483--491.

\bibitem{de2018heterogeneous}
T.~de~Freitas~Pereira, A.~Anjos, and S.~Marcel, ``Heterogeneous face
  recognition using domain specific units,'' \emph{IEEE Transactions on
  Information Forensics and Security}, vol.~14, no.~7, pp. 1803--1816, 2018.

\bibitem{bhatt2012memetically}
H.~S. Bhatt, S.~Bharadwaj, R.~Singh, and M.~Vatsa, ``Memetically optimized
  mcwld for matching sketches with digital face images,'' \emph{IEEE
  Transactions on Information Forensics and Security}, vol.~7, no.~5, pp.
  1522--1535, 2012.

\bibitem{parkhi2015deep}
O.~M. Parkhi, A.~Vedaldi, A.~Zisserman \emph{et~al.}, ``Deep face
  recognition.'' in \emph{bmvc}, vol.~1, no.~3, 2015, p.~6.

\bibitem{wen2016discriminative}
Y.~Wen, K.~Zhang, Z.~Li, and Y.~Qiao, ``A discriminative feature learning
  approach for deep face recognition,'' in \emph{European conference on
  computer vision}.\hskip 1em plus 0.5em minus 0.4em\relax Springer, 2016, pp.
  499--515.

\bibitem{phillips2000feret}
P.~J. Phillips, H.~Moon, S.~A. Rizvi, and P.~J. Rauss, ``The feret evaluation
  methodology for face-recognition algorithms,'' \emph{IEEE Transactions on
  pattern analysis and machine intelligence}, vol.~22, no.~10, pp. 1090--1104,
  2000.

\bibitem{deng2019mutual}
Z.~Deng, X.~Peng, Z.~Li, and Y.~Qiao, ``Mutual component convolutional neural
  networks for heterogeneous face recognition,'' \emph{IEEE Transactions on
  Image Processing}, vol.~28, no.~6, pp. 3102--3114, 2019.

\bibitem{shao2016cross}
M.~Shao and Y.~Fu, ``Cross-modality feature learning through generic
  hierarchical hyperlingual-words,'' \emph{IEEE transactions on neural networks
  and learning systems}, vol.~28, no.~2, pp. 451--463, 2016.

\bibitem{liu2017sphereface}
W.~Liu, Y.~Wen, Z.~Yu, M.~Li, B.~Raj, and L.~Song, ``Sphereface: Deep
  hypersphere embedding for face recognition,'' in \emph{Proceedings of the
  IEEE conference on computer vision and pattern recognition}, 2017, pp.
  212--220.

\bibitem{wang2017normface}
F.~Wang, X.~Xiang, J.~Cheng, and A.~L. Yuille, ``Normface: L2 hypersphere
  embedding for face verification,'' in \emph{Proceedings of the 25th ACM
  international conference on Multimedia}, 2017, pp. 1041--1049.

\end{thebibliography}
}

%
\vspace{-2.5\baselineskip}
\begin{IEEEbiography}[{\includegraphics[width=0.9in,height=1.25in,clip,keepaspectratio]{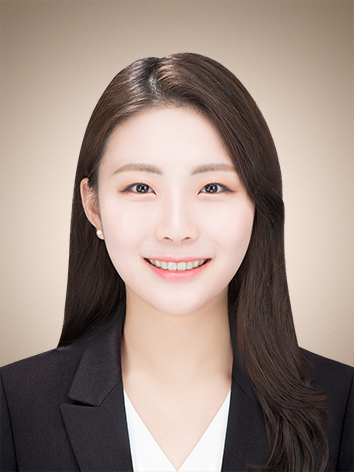}}]{MyeongAh Cho}
received her B.S. degree in Electronic Engineering from Kyunghee University, in 2018. She is currently
pursuing the Ph.D. degree in Electrical and Electronic engineering from Yonsei University, Seoul, South Korea. Her current research interests include heterogeneous face recognition and video recognition via deep learning.
\vspace{-2.8\baselineskip}
\end{IEEEbiography}
\begin{IEEEbiography}[{\includegraphics[width=0.9in,height=1.25in,clip,keepaspectratio]{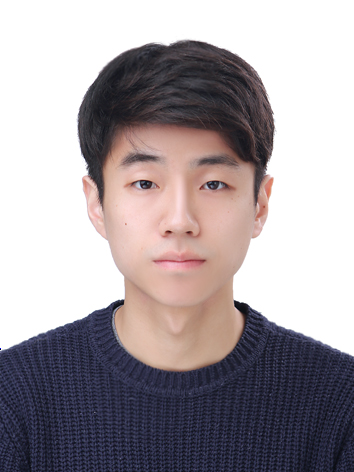}}]{Taeoh Kim}
received his B.S. degree in Electrical and Electronic Engineering from Yonsei University, Seoul, South Korea, in 2015, in where he is currently pursuing the Ph.D. degree. His current research interests include image restoration and computer vision via deep learning.
\vspace{-2.8\baselineskip}
\end{IEEEbiography}

\begin{IEEEbiography}[{\includegraphics[width=0.9in,height=1.25in,clip,keepaspectratio]{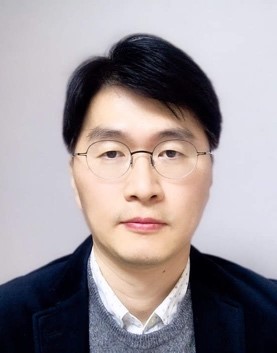}}]{Ig-Jae Kim} is currently a Director of Center for Imaging Media Research, Korea Institute of Science and Technology (KIST), South Korea. He is also an associate professor at Korea University of Science and Technology. He received his Ph.D. degree from EECS of Seoul National University in 2009, M.S. and B.S. degree from EE of Yonsei University, South Korea, in 1998 and 1996 respectively. He had worked in Massachusetts Institute of Technology (MIT) Media Lab as a postdoctoral researcher (2009-2010). 
He is interested in pattern recognition, computer vision and graphics, deep learning, and computational photography.
\vspace{-2.8\baselineskip}
\end{IEEEbiography}

\begin{IEEEbiography}[{\includegraphics[width=0.9in,height=1.25in,clip,keepaspectratio]{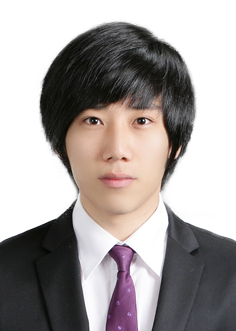}}]{Kyungjae Lee}
is an Assistant Professor of Computer Science at Yongin University, South Korea. He received his Ph.D. and M.S. in Electrical and EE from Yonsei University, Seoul, South Korea, in 2018 and 2013, and his B.S. in Electronics and Radio Engineering from Kyunghee University, South Korea, in 2011. After graduate school, he served as a Staff Software Engineer of the Mobile Communications Business at Samsung Electronics. His research interests include multi-sensor-based computer vision and advanced driver assistance systems.
\vspace{-2.8\baselineskip}
\end{IEEEbiography}

\begin{IEEEbiography}[{\includegraphics[width=0.9in,height=1.25in,clip,keepaspectratio]{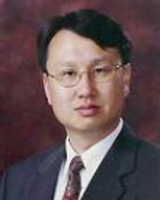}}]{Sangyoun Lee}
(M’04) received his B.S. and M.S. degrees in Electrical and Electronic Engineering from Yonsei University, Seoul, South Korea, in 1987 and 1989, respectively, and his Ph.D. degree in Electrical and Computer Engineering from the Georgia Institute of Technology, Atlanta, GA, USA in 1999. He is currently a Professor of Electrical and Electronic Engineering with the Graduate School, and the Head of the Image and Video Pattern Recognition Laboratory, Yonsei University. His research interests include all aspects of computer vision, with a special focus on pattern recognition for face detection and recognition, advanced driver-assistance systems, and video codecs.
\end{IEEEbiography}




\end{document}